\documentclass[10pt, a4paper, twocolumn]{article}

\usepackage[utf8]{inputenc}
\usepackage[T1]{fontenc}
\usepackage[english]{babel}
\usepackage{tgheros}

\usepackage[varqu]{zi4}
\usepackage{microtype}

\usepackage{titlesec}
\titlespacing*{\section}{0pt}{1.5ex plus 1ex minus .2ex}{0.1em}
\titlespacing*{\subsection}{0pt}{1.5ex plus 1ex minus .2ex}{0.1em}
\titlespacing*{\subsubsection}{0pt}{1.5ex plus 1ex minus .2ex}{0.1em}
\titleformat{\paragraph}[runin]{\normalfont\bfseries}{}{0pt}{}[.]
\titlespacing*{\paragraph}{0pt}{1ex plus .5ex minus .2ex}{0.5em}

\usepackage[table]{xcolor}
\usepackage{booktabs}
\usepackage{float}
\usepackage{graphicx}
\usepackage[labelfont={bf,small}, textfont={small}]{caption}
\definecolor{acqua_celeste}{RGB}{0, 160, 175}
\definecolor{oursrow}{HTML}{DCEFE0}

\usepackage{amsmath}
\usepackage{amssymb}
\usepackage{mathtools}
\usepackage{array}
\usepackage{xurl}

\usepackage[left=1.5cm, right=1.5cm, top=1.5cm, bottom=2cm]{geometry}
\usepackage[numbers, sort&compress]{natbib}
\usepackage{hyperref}
\hypersetup{
    colorlinks=true,
    linkcolor=black,
    urlcolor=acqua_celeste,
    citecolor=acqua_celeste
}

\usepackage{fancyhdr}
\usepackage{authblk}
\newcommand{\ours}{DIAPASON}
\newcommand{\oursfull}{\textbf{DI}stillation \textbf{A}cross \textbf{P}ixels, \textbf{A}ffinities, \textbf{S}cenes, and \textbf{O}utputs from One \textbf{N}etwork (\textbf{\ours{}})}
\newcommand{\reft}{reference}
\newcommand{\Reft}{Reference}
\newcommand{\pmsd}[1]{\ensuremath{_{\pm#1}}}
\DeclareMathOperator{\KL}{KL}
\DeclareMathOperator{\CE}{CE}
\DeclareMathOperator{\clip}{clip}
\DeclareMathOperator{\rnd}{round}

\newcolumntype{L}[1]{>{\raggedright\arraybackslash}p{#1}}
\newsavebox{\fitbox}
\let\plaintabular\tabular \let\endplaintabular\endtabular
\AtBeginDocument{\renewenvironment{tabular}{\begin{lrbox}{\fitbox}\plaintabular}
  {\endplaintabular\end{lrbox}\ifdim\wd\fitbox>\linewidth\resizebox{\linewidth}{!}{\usebox{\fitbox}}\else\usebox{\fitbox}\fi}}

\title{\huge\bfseries\vspace{-1em} DIAPASON: Quantization-Aware Self-Distillation for Classification and Dense Prediction on the GAP9 Microcontroller}

\author[1]{Eugenio Lomurno\textsuperscript{*}}
\author[1]{Greta Corti}
\author[1]{Giacomo Bossi}
\author[2]{Manuel Pariente}
\author[1]{Matteo Matteucci}

\affil[1]{Politecnico di Milano, AIRLab, Italy}
\affil[2]{EssilorLuxottica Italia S.p.A., Italy}

\date{}

\begin{document}

\twocolumn[
  \begin{@twocolumnfalse}
    \maketitle
    \vspace{-3em}

    \begin{abstract}
        \setlength{\parindent}{0pt}
        \setlength{\parskip}{4pt}
        \itshape
        \noindent Neural accelerators can run deep networks on microcontrollers, whose memory is limited. Quantization reduces the memory footprint, and some accelerators accept 2-bit weights. Quantization is applied after or during training. Post-training quantization converts a trained network without retraining it, but can degrade accuracy, especially at low bit-widths. Quantization-aware training simulates quantization during training, so that the network adapts to it. Recent methods distill the quantized network, the student, from its full-precision version, the teacher. This work proposes \oursfull{}, a quantization-aware self-distillation method for classification, segmentation and depth estimation. The student learns from the teacher's outputs and the task labels. In segmentation and depth estimation, distillation covers every pixel and adds pair-wise feature affinities. Segmentation also uses an adversarial term on the whole output map. On the CIFAR-10 and CIFAR-100 classification benchmark of prior self-distillation methods, \ours{} improves the best published result in 9 of 12 configurations. Beyond this benchmark, \ours{} brings its students to the GAP9 microcontroller as fully integer networks, with 8-bit activations and first and last layers. Before export, all these students with 8-bit weights, and the dense ones with 4-bit weights, match or exceed their teachers, whereas post-training quantization loses up to 85\% of the teacher's task metric even at 8 bits. With 2-bit weights, the students keep on average 97.6\% of their teachers' task metric in classification and segmentation, and run up to 2.82 times faster on the GAP9 chip than with 8-bit weights.
    \end{abstract}

    \vspace{0.5em}

    \vspace{1em}
    \hrule height 1pt
    \vspace{2em}
  \end{@twocolumnfalse}
]

{
  \renewcommand{\thefootnote}{\fnsymbol{footnote}}
  \footnotetext[1]{Corresponding author: \texttt{eugenio.lomurno@polimi.it}}
}

\section{Introduction}\label{sec:intro}

Running a deep network on a microcontroller, next to the sensor, avoids sending the data to a remote server: the response is faster, the energy of wireless transmission is saved and the data stay on the device \cite{banbury2021mlperftiny}. The constraint is memory. A microcontroller has from about a hundred kilobytes to a few megabytes \cite{saha2022mcureview}, two to three orders of magnitude less than a mobile phone \cite{lin2020mcunet}. Quantization reduces the footprint of a network by representing its weights and activations with few bits, and some recent microcontrollers include neural accelerators that compute in integer arithmetic with weights of 2 to 8 bits \cite{gap9}. Fewer bits, however, mean a larger rounding error, and a quantized network is useful only if it keeps most of the accuracy of its full-precision version.

Post-training quantization (PTQ) obtains the quantized network from a trained full-precision one by calibrating the quantization ranges on a few batches of data, and works well at 8 bits on many architectures. With fewer bits, and sometimes even at this bit-width, the rounding error costs accuracy. Quantization-aware training (QAT) reduces this drop by simulating quantization during training, so that weights and ranges adapt to the rounding \cite{jacob2018integer,nagel2021whitepaper}. QAT methods differ in how they train the quantized network, for example in how the gradient crosses the rounding \cite{lee2021ewgs}.

Recent QAT methods rely on distillation: a full-precision teacher guides the quantized student, typically by minimizing the Kullback--Leibler divergence between their softened outputs \cite{kim2019qkd}. In quantization-aware self-distillation, the teacher is the full-precision version of the network itself and the student is its quantized copy, initialized from the teacher's weights \cite{zhao2024sqakd}. Later works add a distance between intermediate features of the student and the teacher \cite{lee2026squat}, or select the data augmentation with a metric computed from the teacher \cite{kur2025daqakd}. These methods address classification, usually with the first and last layers in floating point. None of the self-distillation methods is deployed on a microcontroller: at most, they measure the inference speed of 8-bit models on a GPU board \cite{zhao2024sqakd,lee2026squat}.

Quantization has also been applied to dense prediction tasks, such as semantic segmentation and depth estimation, whose output is a map with one prediction per pixel. Segmentation and depth networks have been quantized after training \cite{shen2025quartdepth} or during training \cite{yu2019gdrq,lu2022quantdepth}, and distillation during quantization-aware training has been applied to segmentation \cite{yu2026mfd,zhang2021medq,liu2021zaq}. To our knowledge, however, no quantization-aware self-distillation method addresses classification, segmentation and depth estimation together, with weights down to 2 bits in all three tasks and deployment on a microcontroller.

To address this gap, we propose \oursfull{}, a quantization-aware self-distillation method for classification and dense prediction. The quantized copy of a full-precision network learns from the task loss and from the distillation of the teacher's outputs. In segmentation and depth estimation, this distillation applies to every pixel and is complemented by pair-wise feature affinities and, for segmentation, by an adversarial term on the whole output map. For each task, the same objective is used at every bit-width of weights and activations and in both quantization schemes under which \ours{} is evaluated.

The first is the benchmark scheme of prior work, used for the comparison with the published methods on CIFAR: weights and activations have the same bit-width, and the first and last layers stay in floating point. In this scheme, the results of \ours{} are averaged over three seeds. The second is the device scheme, in which the network runs natively on the NE16 accelerator of the GAP9 microcontroller. The whole network is integer: the weights have 2, 4 or 8 bits, while the activations, the first and last layers, and the input and output of the network have 8 bits. In this scheme, \ours{} is compared with PTQ and with a QAT baseline trained without distillation; the students are compiled with the GAP9 toolchain, and the latency and energy per inference of each setting are measured on the GAP9 chip.

The contributions of this work are:
\begin{itemize}
\item \textbf{One method for three tasks.} \ours{} applies quantization-aware self-distillation to classification, semantic segmentation and depth estimation networks, at the bit-widths chosen for weights and activations.
\item \textbf{Accuracy.} \ours{} improves the best published result on CIFAR-10 and CIFAR-100 in 9 of 12 configurations. Beyond this benchmark, it is more accurate than the QAT baseline in all 21 classification settings and, on average over three seeds, in 15 of the 18 segmentation and depth settings.
\item \textbf{Deployment on a microcontroller.} The students of the three tasks are deployed on GAP9, where 2-bit weights speed up inference up to 2.82 times over 8-bit weights.
\end{itemize}

Section~\ref{sec:related} reviews related work, Section~\ref{sec:method} presents \ours{} and Section~\ref{sec:setup} the experimental setup. Sections~\ref{sec:bench} and~\ref{sec:device} report the results in the benchmark and device schemes, and Section~\ref{sec:gap9} the deployment on GAP9. Section~\ref{sec:discussion} discusses limitations and future directions, and Section~\ref{sec:conclusion} concludes.

\section{Related work}\label{sec:related}

\paragraph{Quantization}
The quantizer of QAT has been parametrized in several ways. DoReFa-Net normalizes the weights with a hyperbolic tangent before uniform quantization \cite{zhou2016dorefa}, PACT learns the clipping value of the activations \cite{choi2018pact}, LSQ learns the quantization step size \cite{esser2020lsq}, and EWGS learns the quantization intervals and scales the gradient of each element according to its quantization error \cite{lee2021ewgs}. Re-implemented in a common framework, QAT methods differ less than in their original reports, and none is best on all architectures \cite{li2021mqbench}. \ours{} keeps the uniform quantizer of EWGS and changes the training objective (Section~\ref{sec:method-quant}).

On dense tasks, QuartDepth quantizes monocular depth networks to 4 bits after training \cite{shen2025quartdepth}, GDRQ quantizes a PSPNet on Cityscapes \cite{cordts2016cityscapes} with distribution reshaping \cite{yu2019gdrq}, and Lu et al. \cite{lu2022quantdepth} fine-tune a self-supervised depth network under quantization, at 4 bits in the encoder and 8 bits in the decoder; none of them uses a teacher.

\paragraph{Knowledge distillation}
Knowledge distillation was developed to compress an ensemble of models into a single model that is easier to deploy: a student is trained to match the softened output distribution of a teacher \cite{hinton2015kd}. How much the student gains depends on the teacher, and a more accurate teacher does not always produce a more accurate student \cite{cho2019efficacy}.

On dense tasks, distillation has usually compressed a large full-precision teacher into a smaller full-precision student \cite{liu2020skd,shu2021cwd}. Structured Knowledge Distillation (SKD) transfers three levels of structure from a large segmentation network to a compact one: the class distribution of each pixel, the similarity between pairs of positions of a feature map, and a holistic score given by an adversarially trained critic \cite{liu2019skd}. Its extended version applies the pair-wise and pixel-wise terms to object detection and the pair-wise and holistic terms to depth estimation \cite{liu2020skd}. Channel-Wise Knowledge Distillation normalizes each channel of a dense map into a spatial distribution and matches it with the Kullback--Leibler divergence \cite{shu2021cwd}. Teacher and student share the architecture in some depth estimation work, to estimate uncertainty \cite{poggi2020uncertainty} or to improve accuracy \cite{he2026distillanydepth}, not to compress the network.

In quantization-aware training, Apprentice \cite{mishra2018apprentice} and QKD \cite{kim2019qkd} distill a low-precision student from a full-precision teacher, PTG trains the quantized network in progressive stages \cite{zhuang2018ptg}, SPEQ builds the teacher from the student itself with stochastic precision \cite{boo2021speq}, and CMT-KD uses several quantized teachers \cite{pham2023cmtkd}; most of these methods need several training phases or train the teacher jointly with the student \cite{zhao2024sqakd}. Self-Supervised Quantization-Aware Knowledge Distillation (SQAKD) drops both requirements: the teacher is the pretrained full-precision copy of the student, the student starts from the teacher weights, and the loss is the Kullback--Leibler divergence alone, without labels or balancing weights \cite{zhao2024sqakd}. SQuaT adds the distance between an intermediate feature of the student and the teacher feature projected onto the quantization grid of the student \cite{lee2026squat}. Both methods use the EWGS quantizer for convolutional networks on CIFAR and measure inference at 8 bits on an NVIDIA Jetson Nano, a board with a GPU \cite{zhao2024sqakd,lee2026squat}. Data-Augmented Quantization-Aware Knowledge Distillation (DA-QAKD) ranks candidate data augmentations with a metric computed from the teacher, without training the student, and adds cross-entropy to the distillation term \cite{kur2025daqakd}. None of these methods is evaluated on semantic segmentation or depth estimation.

Distillation during quantization-aware training has also been applied to segmentation. MFD distills DeepLabV3, PSPNet and SegFormer students with weights down to 2 bits from a larger teacher whose quantization parameters are perturbed \cite{yu2026mfd}, and ZAQ distills a quantized DeepLabV3 from its full-precision version on synthetic images instead of real data \cite{liu2021zaq}. Other works address a single domain: medical images with 1- to 4-bit networks \cite{zhang2021medq,wang2026q2} and crop and weed segmentation with a 1-bit network \cite{khan2025biagrinet}. GPUSQ-ViT distills 8- and 4-bit transformers from a sparse copy of the same network in classification, detection and segmentation, but does not address depth estimation \cite{yu2023gpusqvit}. These works measure runtime on GPUs or CPUs \cite{yu2026mfd,zhang2021medq,yu2023gpusqvit,khan2025biagrinet}; Nano-U runs on an ESP32-S3 microcontroller, for one binary segmentation task at 8 bits and with a much larger teacher \cite{pizzolato2026nanou}.

On monocular depth, the works reviewed here that use a full-precision reference do not train the quantized weights. QSCA adds auxiliary modules in 16-bit floating point to a 4-bit network and trains only these modules on unlabeled images, with the full-precision copy as the teacher \cite{yang2025qsca}, and SPA-Q fits channel-wise affine parameters to the full-precision features and depth maps after quantization \cite{choi2026spaq}.

\ours{} extends the self-distillation of SQAKD to semantic segmentation and monocular depth estimation with the structured terms of SKD, and trains students with weights down to 2 bits in all three tasks.

\paragraph{Deployment on microcontrollers}
Rusci et al. \cite{rusci2020mixed} lower the most memory-demanding layers of a classification network to 4 or 2 bits until it fits the memory of an STM32H7 microcontroller, and convert it to integer-only form after quantization-aware training. Microcontrollers with a neural accelerator differ in the precision they accept. The eIQ Neutron of the NXP MCX N series uses 8-bit weights \cite{millar2025munpu}, whereas the Arm Ethos-U accelerators use 8-bit weights and, for selected operators, also accept 4-bit weights through the TOSA compilation path \cite{armvela}. The MAX78000 accepts weights of 1, 2, 4 or 8 bits \cite{adimax78000}, the F28P55x of Texas Instruments 2, 4 or 8 bits \cite{tinpuguide}, the STM32N6 4 or 8 bits \cite{stneuralart} and the NE16 accelerator of GAP9 any width from 2 to 8 bits \cite{gap9,ne16repo}, in all cases with 8-bit activations available. A comparison of microcontroller accelerators uses 8-bit models only and leaves accuracy out of its scope \cite{millar2025munpu}. On GAP9, which combines a cluster of RISC-V cores with NE16 \cite{gap9}, this work deploys students with 2-, 4- and 8-bit weights and reports their task metric together with the latency and energy measured on the chip.

\section{Method}\label{sec:method}

\begin{figure*}[tbp]
\centering
\includegraphics[width=\linewidth]{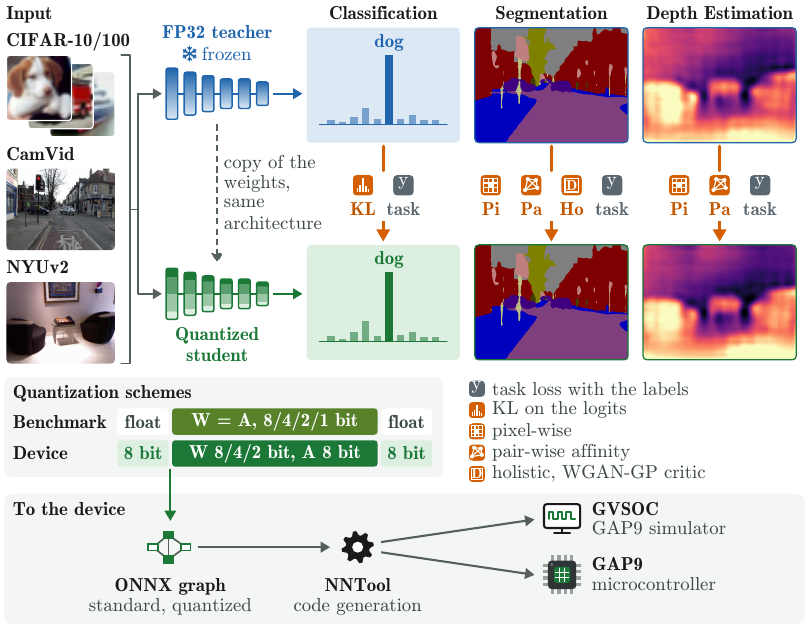}
\caption[Overview of \ours{}]{\textbf{Overview of \ours{}.} Top: the frozen FP32 teacher and its quantized students in classification, segmentation and depth estimation, with the objective terms of each task. Bottom: the benchmark and device schemes, and the deployment pipeline on GAP9.}
\label{fig:overview}
\end{figure*}

\ours{} is the quantization-aware training method proposed in this work. The teacher is a full-precision network trained for the task; the student is its quantized copy, which starts from the teacher's weights and is trained with the task loss and with distillation from the teacher (Figure~\ref{fig:overview}). The method addresses classification, semantic segmentation and depth estimation, with one objective per task (Sections~\ref{sec:method-cls} and~\ref{sec:method-dense}). The bit-widths of weights and activations and the precision of the first and last layers are set independently (Section~\ref{sec:method-quant}), and the objective of each task does not change with them. After training, the students of the device scheme are exported and run on GAP9 (Section~\ref{sec:method-export}).

\subsection{Quantizer}\label{sec:method-quant}
\ours{} quantizes weights and activations with the uniform quantizer of EWGS \cite{lee2021ewgs}. For a layer with weights $w$ and input activations $a$, the quantizer clips $w$ to a learnable range $[l_W,u_W]$ and $a$ to a learnable range $[l_A,u_A]$, normalizes both to $[0,1]$, and rounds them to $2^{b_W}$ and $2^{b_A}$ levels, where $b_W$ and $b_A$ are the bit-widths of weights and activations; the rounded weights are then mapped to $[-1,1]$:
\begin{equation}
\begin{gathered}
\hat w = 2\,Q_{b_W}\!\left(\clip\!\Big(\frac{w-l_W}{u_W-l_W},0,1\Big)\right)-1,\\
\hat a = Q_{b_A}\!\left(\clip\!\Big(\frac{a-l_A}{u_A-l_A},0,1\Big)\right),\\
Q_b(t)=\frac{\rnd\big((2^b-1)\,t\big)}{2^b-1}.
\end{gathered}
\label{eq:quant}
\end{equation}
The layer then applies its convolution or matrix product to $\hat w$ and $\hat a$ and multiplies the output by a learned output scale $s$. Each range is shared by a whole tensor, except for the weights of depthwise convolutions and of the linear layers of DeiT-Tiny, whose output channels differ most in range and therefore have one range each (\ref{app:setups}). The ranges and output scales of all layers form the quantizer parameters $\Lambda$, and $f_{\theta,\Lambda}$ denotes the quantized network with weights $\theta$.

In the backward pass, the rounding has zero gradient almost everywhere. EWGS lets the gradient cross it and scales each element according to its rounding error, with a factor set from an estimate of the Hessian \cite{lee2021ewgs}; the papers of SQAKD and SQuaT describe a nonzero factor for their convolutional networks on CIFAR \cite{zhao2024sqakd,lee2026squat}. \ours{} sets this factor to zero in every network, so that the gradient crosses the rounding unchanged, as with the straight-through estimator \cite{bengio2013ste}.

Before training, the ranges are initialized from percentiles. For the activations, $l_A$ and $u_A$ are the 0.1th and 99.9th percentiles of the layer input, averaged over 8 calibration batches. The weight range is symmetric, with $u_W=-l_W$ equal to the 99.9th percentile of $|w|$. The output scale starts at $s=u_W(u_A-l_A)$, which maps the product of the normalized weights and activations back to the range of the float layer when $l_A=0$; for a nonzero $l_A$, the shift by $l_A$ changes each output channel by a constant, which is folded into the bias at initialization, where the layer has one, so that the student starts close to the teacher. The public code of SQAKD initializes the ranges on one batch, from three standard deviations, and the output scale from the ratio between the mean absolute outputs of the float and quantized layer \cite{zhao2024sqakd}. \ours{} uses this one-batch initialization at 1 bit, and custom ranges for some DeiT-Tiny students (\ref{app:setups}).

The first and last layers can stay in floating point or be quantized with the same quantizer, with a bit-width of their own; Section~\ref{sec:setup} defines the two quantization schemes of the experiments.

\subsection{Self-distillation setting and classification objective}\label{sec:method-cls}
In all three tasks, \ours{} follows the self-distillation setting of SQAKD \cite{zhao2024sqakd}. The teacher $f_{\theta^*}$ is the trained full-precision network, with weights $\theta^*$, and stays frozen; the student $f_{\theta,\Lambda}$ is its quantized copy, with $\theta$ initialized to $\theta^*$. Training minimizes a loss $\mathcal{L}(\theta,\Lambda)$ over both the weights and the quantizer parameters of the student. For classification, the loss adds the cross-entropy with the label to a distillation term that matches the softened output distributions of student and teacher:
\begin{equation}
\begin{gathered}
\mathcal{L}_{\mathrm{cls}} = \mathcal{L}_{\mathrm{CE}} + \mathcal{L}_{\mathrm{KL}},\qquad \mathcal{L}_{\mathrm{CE}}=\CE(z^s,y),\\
\mathcal{L}_{\mathrm{KL}}=T^2\,\KL\!\big(\sigma(z^t/T)\,\big\|\,\sigma(z^s/T)\big),\qquad T=4,
\end{gathered}
\label{eq:cls}
\end{equation}
where $z^s$ and $z^t$ are the logits of student and teacher, $y$ is the label, $\sigma$ is the softmax, $\CE$ and $\KL$ are the cross-entropy and the Kullback--Leibler divergence, and $T$ is the temperature, set to 4 as in SQAKD \cite{zhao2024sqakd}. The factor $T^2$ compensates the $1/T^2$ scaling of the gradients of the softened outputs \cite{hinton2015kd}. SQAKD minimizes $\mathcal{L}_{\mathrm{KL}}$ alone, without labels \cite{zhao2024sqakd}, and DA-QAKD gives $\mathcal{L}_{\mathrm{KL}}$ twice the weight of $\mathcal{L}_{\mathrm{CE}}$ \cite{kur2025daqakd}. \ours{} weights the two terms equally for every network and bit-width; \ref{app:ablations} compares this objective, under the same teacher, with $\mathcal{L}_{\mathrm{KL}}$ alone, $\mathcal{L}_{\mathrm{CE}}$ alone and $\mathcal{L}_{\mathrm{KL}}$ with double weight.

\subsection{Structured distillation for dense prediction}\label{sec:method-dense}
Following SKD \cite{liu2019skd,liu2020skd}, the distillation signal for dense prediction acts at three levels: on single output pixels, with the pixel-wise term $\mathcal{L}_{\mathrm{pi}}$; on pairs of positions of an intermediate feature map, with the pair-wise term $\mathcal{L}_{\mathrm{pa}}$; and, in segmentation only, on the whole output map, with the adversarial holistic term $\mathcal{L}_{\mathrm{ho}}$. The objectives of the two tasks are:
\begin{equation}
\begin{gathered}
\mathcal{L}_{\mathrm{seg}} = \mathcal{L}_{\mathrm{CE}} + \mathcal{L}_{\mathrm{pi}} + \mathcal{L}_{\mathrm{pa}} + \lambda_{\mathrm{ho}}\,\mathcal{L}_{\mathrm{ho}},\qquad \lambda_{\mathrm{ho}}=0.1,\\
\mathcal{L}_{\mathrm{depth}} = \mathcal{L}_{\mathrm{berHu}} + \mathcal{L}_{\mathrm{pi}} + \mathcal{L}_{\mathrm{pa}}.
\end{gathered}
\label{eq:dense}
\end{equation}
In segmentation, $\mathcal{L}_{\mathrm{CE}}$ is the cross-entropy with the label of each pixel, with void pixels ignored; in depth estimation, the task loss $\mathcal{L}_{\mathrm{berHu}}$ is the reverse Huber loss with the ground-truth depth \cite{laina2016berhu}. SKD weights the pixel-wise and pair-wise terms by 10 and the holistic term by 0.1 \cite{liu2020skd}; \ours{} gives weight 1 to the pixel-wise and pair-wise terms and keeps 0.1 for the holistic term, for all networks and bit-widths.

\paragraph{Pixel-wise term}
With $\Omega$ the set of output pixels, $z^s_p$ and $z^t_p$ the class logits of student and teacher at pixel $p$ in segmentation, and $d^s_p$ and $d^t_p$ their predicted depths, the pixel-wise term applies $\mathcal{L}_{\mathrm{KL}}$ of Eq.~\eqref{eq:cls} to every pixel in segmentation and is the mean absolute difference from the teacher's depth in depth estimation:
\begin{equation}
\begin{gathered}
\mathcal{L}_{\mathrm{pi}}^{\mathrm{seg}} = \frac{T^2}{|\Omega|}\sum_{p\in\Omega}\KL\!\big(\sigma(z^t_p/T)\,\big\|\,\sigma(z^s_p/T)\big),\\
\mathcal{L}_{\mathrm{pi}}^{\mathrm{depth}} = \frac{1}{|\Omega|}\sum_{p\in\Omega}\big|d^s_p-d^t_p\big|.
\end{gathered}
\label{eq:pi}
\end{equation}
SKD casts depth estimation as classification over discrete depth intervals and uses the pair-wise and holistic terms without the pixel-wise one \cite{liu2020skd}; \ours{} treats depth as a regression, matches the teacher's depth directly and uses no holistic term.

\paragraph{Pair-wise term}
The pair-wise term trains the student to reproduce how similar the positions of an intermediate feature map are to one another in the teacher. The feature map is taken at the same layer of both networks: the output of the pyramid pooling module for Fast-SCNN and R18-PSPNet, and the bottleneck for UNet-small. It is average-pooled to a square grid of $\min(h,16)\times\min(h,16)$ cells, where $h$ is its height, so that it has $N=\min(h,16)^2$ positions (\ref{app:setups} gives $N$ for each network). The affinity $a_{ij}$ between positions $i$ and $j$ is the cosine similarity of their feature vectors $f_i$ and $f_j$, and the pair-wise term is the mean squared difference between the affinities of student and teacher:
\begin{equation}
\begin{gathered}
a_{ij} = \frac{f_i^{\top} f_j}{\|f_i\|_2\,\|f_j\|_2},\\
\mathcal{L}_{\mathrm{pa}} = \frac{1}{N^2}\sum_{i=1}^{N}\sum_{j=1}^{N}\big(a^s_{ij}-a^t_{ij}\big)^2 .
\end{gathered}
\label{eq:pa}
\end{equation}
This is the pair-wise loss of SKD computed over all pairs of positions \cite{liu2020skd}. Because teacher and student share the architecture, the two feature maps have the same shape and need no adaptation layer.

\paragraph{Holistic term}
In segmentation, a critic network $D_\phi$ learns to tell the output map of the teacher from that of the student, given the input image, and the student is trained to raise the score that the critic gives to its map. The input of the critic is $u=[x;\sigma(z)]$, the input image $x$ concatenated along the channels with the map of class probabilities, computed from the logits of the student ($u^s$) or of the teacher ($u^t$). As in SKD \cite{liu2020skd}, the critic is trained with the Wasserstein objective and a gradient penalty \cite{gulrajani2017wgangp}:
\begin{equation}
\begin{aligned}
&\max_{\phi}\;\mathbb{E}\big[D_\phi(u^t)\big]-\mathbb{E}\big[D_\phi(u^s)\big]\\
&\qquad-\lambda_{\mathrm{gp}}\,\mathbb{E}\Big[\big(\|\nabla_{\hat u}D_\phi(\hat u)\|_2-1\big)^2\Big],\\
&\mathcal{L}_{\mathrm{ho}}=-\mathbb{E}\big[D_\phi(u^s)\big],
\end{aligned}
\label{eq:ho}
\end{equation}
where $\hat u$ is a random interpolation between $u^t$ and $u^s$ and $\lambda_{\mathrm{gp}}=10$. Critic and student are updated in alternation, with one critic step before each student step; the critic is used only during training.

\subsection{Deployment pipeline}\label{sec:method-export}
After training, each student of the device scheme is exported to an ONNX graph in the QDQ format, in which QuantizeLinear and DequantizeLinear nodes placed around standard operators represent the quantization \cite{onnxruntime}. The weights keep the integer codes of training, with one scale per output channel once batch normalization is folded in (\ref{app:setups}), and each activation range becomes a scale and an 8-bit zero-point. Each graph is then checked on two conditions. Its structure must match the graph that ONNX Runtime static quantization \cite{onnxruntime} produces from the same float model, the only difference being the number of weight levels when the weights have fewer than 8 bits. Its task metric in ONNX Runtime must stay close to that of the trained student at the same input resolution: it may drop by at most 2.5 points of top-1 accuracy or 1.5 mIoU, and the RMSE may change by at most 3\%. The checked graph then goes through the GAP9 developer flow: NNTool imports it and generates code that runs either on the chip or on GVSOC, a simulator of GAP9 that executes the same code \cite{gapsdknntool,gapquantimport,bruschi2021gvsoc}. Section~\ref{sec:gap9} reports the task metric of the exported graphs, measured in ONNX Runtime and GVSOC, and the latency and energy per inference measured on the chip.

\section{Experimental setup}\label{sec:setup}

\paragraph{Quantization schemes}
In the \emph{benchmark scheme}, $b_W=b_A\in\{1,2,4,8\}$, and the first and last layers stay in floating point; its settings are labeled W1A1, W2A2, W4A4 and W8A8. In the \emph{device scheme}, which matches the integer pipeline of GAP9, $b_W\in\{2,4,8\}$, $b_A=8$, and the first and last layers are quantized to 8 bits, so that the whole graph is integer; its settings are labeled W8A8, W4A8 and W2A8. During training, the output of the last layer is not quantized; it receives its 8-bit scale at export.

\paragraph{Datasets and networks}
Classification is evaluated with top-1 accuracy on CIFAR-10, with ResNet-20 and VGG-8, and on CIFAR-100, with ResNet-32 and VGG-13, the benchmarks of SQAKD \cite{zhao2024sqakd,krizhevsky2009cifar,he2016resnet,simonyan2015vgg}, and on Tiny-ImageNet, with ResNet-18, MobileNetV2 and the distilled DeiT-Tiny \cite{le2015tinyimagenet,sandler2018mobilenetv2,touvron2021deit}. Segmentation on CamVid \cite{brostow2009camvid} is evaluated with the mean intersection over union (mIoU), and depth estimation on NYUv2 \cite{silberman2012nyu} with the root mean square error (RMSE, in meters). Both tasks use three networks: R18-PSPNet, a PSPNet head \cite{zhao2017pspnet} on a dilated ResNet-18; Fast-SCNN \cite{poudel2019fastscnn}; and UNet-small, a small U-Net \cite{ronneberger2015unet}. \ref{app:setups} gives the number of test images and the input resolution of each dataset, and the number of parameters of each network.

\paragraph{Teachers}
On CIFAR, each network has two teachers, so that the comparison with published results can separate, as far as possible, the effect of \ours{} from that of a more accurate teacher. The \emph{\reft{} teachers} have an accuracy close to that of the teachers of SQAKD and serve the comparison with SQAKD and SQuaT: on CIFAR-100 they follow the training setup of the public SQAKD code, on CIFAR-10 an SGD setup of ours, and their accuracy is within 0.45 points of the teachers of SQAKD, except for VGG-8 (+0.91; Table~\ref{tab:cifar}). The \emph{final teachers} are more accurate and place \ours{} in the setting of DA-QAKD, whose teachers are trained with CutMix: they are trained with AutoAugment \cite{cubuk2019autoaugment}, Cutout \cite{devries2017cutout} and an exponential moving average of the weights. The Tiny-ImageNet teachers are fine-tuned from ImageNet-1k weights \cite{deng2009imagenet}. The dense teachers are trained on CamVid and NYUv2, starting from models trained on Cityscapes \cite{cordts2016cityscapes}, from an ImageNet-1k backbone or from scratch, depending on the network. \ref{app:setups} gives every teacher setup.

\paragraph{Student training}
On CIFAR, the batch size, weight decay and epochs of the students follow SQAKD \cite{zhao2024sqakd}. From W2A2 to W8A8 and in the device scheme, the CIFAR students add TrivialAugment \cite{muller2021trivialaugment}, the augmentation that DA-QAKD selects for VGG-8 and VGG-13 on CIFAR \cite{kur2025daqakd}, to random crops and flips; at W1A1 they use crops and flips only. The weight learning rate of the dense students is three times higher at W2A8 than at W4A8 and W8A8, and DeiT-Tiny at W2A2 uses a learning rate ten times lower than the other Tiny-ImageNet students (Section~\ref{sec:bench-tiny}). \ref{app:setups} lists the hyper-parameters per task, bit-width and quantization scheme.

\paragraph{Baselines}
The published results are reported as they appear in the cited papers, with their own teachers and training setups; \ref{app:setups} lists the source of each value. The QAT baseline trains the same student as \ours{}, from the same teacher weights and with the same quantizer, optimizer, budget and augmentation, with the task loss only, so that the comparison isolates the contribution of distillation. The PTQ baseline initializes the quantizer as in Section~\ref{sec:method-quant} and updates the batch-normalization statistics, without training.

\paragraph{Evaluation protocol}
Students train for the full budget. PTQ does not train and is evaluated once per calibration seed. Each setting is trained with three seeds, and the tables report the mean and the sample standard deviation; SQAKD and DA-QAKD report one value per setting, and SQuaT does not state its number of runs. Differences between two of our rows are paired by seed, and $k/3$ counts the seeds in which the first of the two is better. Differences and gains are computed before rounding. In classification, the two runs of a pair share the seed, the initial weights, the calibration batches and the seeds of the augmentation workers. In the dense tasks the augmentation is not seeded: the two runs share the initial weights and the images of the calibration batches, but their augmentation differs.

\paragraph{Hardware and software}
Training ran on two servers, with two NVIDIA L40S GPUs (PyTorch 2.12.0, CUDA 13.2) or two NVIDIA RTX A6000 GPUs (PyTorch 2.10.0, CUDA 12.8); \ref{app:setups} lists the runs on each. Section~\ref{sec:gap9} gives the deployment setup.

\section{Results in the benchmark scheme}\label{sec:bench}

\subsection{CIFAR}\label{sec:bench-cifar}
\paragraph{With the \reft{} teachers}
On CIFAR (Table~\ref{tab:cifar}), with teachers of accuracy similar to those of SQAKD, \ours{} is above the best of SQAKD and SQuaT in all eight W2A2 and W4A4 settings, by 1.12 points on average and by 0.19 (ResNet-32, W2A2) to 3.19 points (VGG-13, W4A4). At W1A1, which uses its own training setup (\ref{app:setups}), the comparison is mixed: \ours{} matches SQAKD on VGG-8 (+0.01 points), is below SQuaT by 0.45 (ResNet-20) and 1.05 points (ResNet-32), and is below SQAKD by 1.15 points (VGG-13).

\paragraph{With the final teachers}
The final teachers are 1.9 to 5.1 points more accurate than the teachers of SQAKD and 0.4 to 2.4 points more accurate than the CutMix teachers of DA-QAKD (FP32 rows of Table~\ref{tab:cifar}). With them, \ours{} gives the best result in 9 of the 12 settings, including W1A1 on VGG-8 and VGG-13, by 0.91 and 2.06 points over the best published result; at W4A4 on ResNet-20, its margin over DA-QAKD, 0.02 points, is within one standard deviation of \ours{}. The published results are best in the other three: SQuaT at W1A1 on ResNet-20 and ResNet-32, by 0.77 and 0.30 points, and DA-QAKD at W4A4 on VGG-8, by 0.19 points.

\paragraph{Variability and effect of the teacher}
The results of \ours{} vary little across seeds: in the 32 cells of the \reft{} and final columns of Table~\ref{tab:cifar}, the standard deviation is at most 0.29 points. With the same loss, quantizer and augmentation, the final teacher gives a better student than the \reft{} teacher in all 12 settings from W2A2 to W8A8 (\ref{app:teacher}).

\begin{table*}[tbp]
\centering\scriptsize\setlength{\tabcolsep}{2.2pt}
\caption[CIFAR results in the benchmark scheme]{\textbf{CIFAR, benchmark scheme.} Top-1 accuracy (\%) of the published methods, as reported by each work, and of \ours{} with the \reft{} (ref.)\ and the final teacher, with the first and last layers in floating point, mean and standard deviation over three seeds. FP32: full-precision network or teacher of each work. $\Delta$: \ours{} final minus the best published value. Bold: best in each row.}
\label{tab:cifar}
\begin{tabular}{@{}llcccccccccc@{}}
\toprule
 & & QAT & \multicolumn{3}{c}{QAT with distillation} & \multicolumn{3}{c}{Self-distillation} & \multicolumn{2}{c}{\ours{}} & \\
\cmidrule(lr){3-3}\cmidrule(lr){4-6}\cmidrule(lr){7-9}\cmidrule(lr){10-11}
Network & Bits & EWGS & KD & QKD & SPEQ & SQAKD & SQuaT & DA-QAKD & ref. & final & $\Delta$ \\
 & & \cite{lee2021ewgs} & \cite{zhao2024sqakd} & \cite{kim2019qkd} & \cite{boo2021speq} & \cite{zhao2024sqakd} & \cite{lee2026squat} & \cite{kur2025daqakd} &  &  & \\
\midrule
\multicolumn{12}{@{}l}{\textbf{(a) CIFAR-10}}\\
ResNet-20 & \textcolor{black!55}{FP32} & \textcolor{black!55}{91.40} & \textcolor{black!55}{92.58} & \textcolor{black!55}{91.60} & \textcolor{black!55}{92.10} & \textcolor{black!55}{92.58} & \textcolor{black!55}{92.63} & \textcolor{black!55}{94.00} & \textcolor{black!55}{92.59} & \textcolor{black!55}{94.47} & \\
 & W1A1 & 85.60 & -- & -- & -- & 86.47 & \textbf{86.97}\pmsd{0.14} & -- & 86.52\pmsd{0.26} & 86.20\pmsd{0.21} & $-$0.77 \\
 & W2A2 & -- & 91.44 & 90.50 & 91.40 & 91.80 & 91.90\pmsd{0.11} & 92.33 & 92.23\pmsd{0.07} & \textbf{92.57}\pmsd{0.09} & +0.24 \\
 & W4A4 & -- & -- & 93.10 & -- & 92.59 & 92.67\pmsd{0.03} & 94.39 & 92.92\pmsd{0.04} & \textbf{94.41}\pmsd{0.08} & +0.02 \\
 & W8A8 & -- & -- & -- & -- & -- & -- & -- & 93.05\pmsd{0.05} & \textbf{94.54}\pmsd{0.03} & -- \\
\addlinespace[3pt]
VGG-8 & \textcolor{black!55}{FP32} & \textcolor{black!55}{--} & \textcolor{black!55}{--} & \textcolor{black!55}{--} & \textcolor{black!55}{--} & \textcolor{black!55}{91.27} & \textcolor{black!55}{--} & \textcolor{black!55}{92.91} & \textcolor{black!55}{92.18} & \textcolor{black!55}{93.33} & \\
 & W1A1 & -- & -- & -- & -- & 89.05 & -- & -- & 89.06\pmsd{0.11} & \textbf{89.96}\pmsd{0.14} & +0.91 \\
 & W2A2 & -- & -- & -- & -- & 91.55 & -- & 92.66 & 92.45\pmsd{0.14} & \textbf{92.85}\pmsd{0.06} & +0.19 \\
 & W4A4 & -- & -- & -- & -- & 91.31 & -- & \textbf{93.66} & 92.81\pmsd{0.03} & 93.47\pmsd{0.03} & $-$0.19 \\
 & W8A8 & -- & -- & -- & -- & -- & -- & -- & 92.84\pmsd{0.05} & \textbf{93.64}\pmsd{0.12} & -- \\
\midrule
\multicolumn{12}{@{}l}{\textbf{(b) CIFAR-100}}\\
ResNet-32 & \textcolor{black!55}{FP32} & \textcolor{black!55}{--} & \textcolor{black!55}{--} & \textcolor{black!55}{70.80} & \textcolor{black!55}{70.30} & \textcolor{black!55}{71.33} & \textcolor{black!55}{71.40} & \textcolor{black!55}{74.57} & \textcolor{black!55}{71.69} & \textcolor{black!55}{76.47} & \\
 & W1A1 & -- & -- & -- & -- & 59.41 & \textbf{61.12}\pmsd{0.19} & -- & 60.07\pmsd{0.29} & 60.82\pmsd{0.09} & $-$0.30 \\
 & W2A2 & -- & -- & 66.40 & 69.10 & 69.99 & 70.57\pmsd{0.04} & 69.27 & 70.76\pmsd{0.15} & \textbf{71.25}\pmsd{0.26} & +0.68 \\
 & W4A4 & -- & -- & 73.30 & -- & 71.65 & 71.94\pmsd{0.05} & 75.31 & 72.81\pmsd{0.09} & \textbf{76.35}\pmsd{0.28} & +1.04 \\
 & W8A8 & -- & -- & -- & -- & -- & -- & -- & 72.86\pmsd{0.03} & \textbf{76.93}\pmsd{0.08} & -- \\
\addlinespace[3pt]
VGG-13 & \textcolor{black!55}{FP32} & \textcolor{black!55}{--} & \textcolor{black!55}{76.36} & \textcolor{black!55}{--} & \textcolor{black!55}{--} & \textcolor{black!55}{76.36} & \textcolor{black!55}{--} & \textcolor{black!55}{76.24} & \textcolor{black!55}{76.81} & \textcolor{black!55}{78.62} & \\
 & W1A1 & -- & 68.47 & -- & -- & 68.56 & -- & -- & 67.41\pmsd{0.21} & \textbf{70.62}\pmsd{0.20} & +2.06 \\
 & W2A2 & -- & -- & -- & -- & 74.65 & -- & 76.59 & 76.41\pmsd{0.23} & \textbf{77.29}\pmsd{0.05} & +0.70 \\
 & W4A4 & -- & -- & -- & -- & 74.67 & -- & 77.97 & 77.86\pmsd{0.04} & \textbf{79.18}\pmsd{0.07} & +1.21 \\
 & W8A8 & -- & -- & -- & -- & -- & -- & -- & 78.02\pmsd{0.07} & \textbf{79.26}\pmsd{0.13} & -- \\
\midrule
\multicolumn{11}{@{}l}{Mean $\Delta$ over the 12 settings with published values; $\Delta>0$ in 9 of 12} & +0.48 \\
\bottomrule
\end{tabular}
\end{table*}

\subsection{Tiny-ImageNet}\label{sec:bench-tiny}
\paragraph{Comparison with QAT and PTQ}
On Tiny-ImageNet (Table~\ref{tab:tiny}), \ours{} is above QAT in all nine settings, in 3/3 seeds each, by 2.19 to 17.45 points, and the 95\% confidence interval of the paired difference is above zero in all nine (\ref{app:stats}). It is also above PTQ in all nine settings, by at least 0.24 points (DeiT-Tiny, W8A8). The QAT baseline, trained with the learning rate of \ours{}, is below PTQ in 5 of the 9 settings.

\paragraph{Distance from the teacher}
At W8A8 and W4A4, the student is within 1.65 points of its teacher. At W2A2, the drop from the teacher depends on the network: 4.53 points for ResNet-18, 12.63 for DeiT-Tiny and 19.60 for MobileNetV2. DeiT-Tiny at W2A2 does not converge with the setup of the other settings, so QAT and \ours{} use their own setup there, with, among other changes, a learning rate ten times lower (\ref{app:setups}), and PTQ is evaluated with two calibration seeds.

\paragraph{Published results}
The Tiny-ImageNet results of SQAKD and DA-QAKD on ResNet-18 and MobileNetV2, and those of SQuaT on ResNet-18, use teachers of 65.55 to 66.87\% and 58.07 to 58.64\%, respectively \cite{zhao2024sqakd,lee2026squat,kur2025daqakd}. Our teachers are fine-tuned from ImageNet-1k weights, so they are pretrained on additional data, and are 4 to 13 points more accurate; for this reason, Table~\ref{tab:tiny} does not include these results.

\begin{table*}[tbp]
\centering\scriptsize\setlength{\tabcolsep}{1.8pt}
\caption[Tiny-ImageNet results in the benchmark scheme]{\textbf{Tiny-ImageNet, benchmark scheme.} Top-1 accuracy (\%) of PTQ, QAT and \ours{} with the first and last layers in floating point, mean and standard deviation over three seeds (two calibration seeds for PTQ on DeiT-Tiny at W2A2). Gain over QAT: mean paired difference (\ref{app:stats}). Bold: best student in each column.}
\label{tab:tiny}
\begin{tabular}{@{}lccccccccc@{}}
\toprule
 & \multicolumn{3}{c}{ResNet-18} & \multicolumn{3}{c}{MobileNetV2} & \multicolumn{3}{c}{DeiT-Tiny} \\
 & \multicolumn{3}{c}{\scriptsize FP32 teacher: 71.10} & \multicolumn{3}{c}{\scriptsize FP32 teacher: 71.45} & \multicolumn{3}{c}{\scriptsize FP32 teacher: 73.49} \\
\cmidrule(lr){2-4}\cmidrule(lr){5-7}\cmidrule(lr){8-10}
Method & W8A8 & W4A4 & W2A2 & W8A8 & W4A4 & W2A2 & W8A8 & W4A4 & W2A2 \\
\midrule
PTQ & 68.83\pmsd{0.03} & 67.39\pmsd{0.13} & 4.65\pmsd{0.22} & 68.79\pmsd{0.03} & 59.15\pmsd{0.87} & 0.64\pmsd{0.03} & 73.09\pmsd{0.09} & 64.34\pmsd{0.06} & 0.48\pmsd{0.01} \\
QAT & 65.23\pmsd{0.13} & 64.51\pmsd{0.23} & 56.89\pmsd{0.38} & 67.24\pmsd{0.22} & 65.24\pmsd{0.44} & 34.40\pmsd{1.52} & 62.95\pmsd{0.17} & 57.83\pmsd{0.26} & 58.67\pmsd{0.13} \\
\rowcolor{oursrow}\ours{} & \textbf{70.89}\pmsd{0.23} & \textbf{70.24}\pmsd{0.10} & \textbf{66.57}\pmsd{0.08} & \textbf{71.73}\pmsd{0.03} & \textbf{69.80}\pmsd{0.22} & \textbf{51.85}\pmsd{0.42} & \textbf{73.33}\pmsd{0.05} & \textbf{71.98}\pmsd{0.08} & \textbf{60.86}\pmsd{0.22} \\
\quad Gain over QAT & {\scriptsize +5.66} & {\scriptsize +5.73} & {\scriptsize +9.68} & {\scriptsize +4.49} & {\scriptsize +4.55} & {\scriptsize +17.45} & {\scriptsize +10.37} & {\scriptsize +14.15} & {\scriptsize +2.19} \\
\bottomrule
\end{tabular}
\end{table*}

\section{Results in the device scheme}\label{sec:device}

\subsection{Classification}\label{sec:device-cls}
In the device scheme, all classification students start from the final teacher (Table~\ref{tab:cls-device}). The results of this section are those of the fake-quantized PyTorch models, before export; Section~\ref{sec:gap9} follows them through deployment.

\paragraph{Comparison with QAT and PTQ}
\ours{} is above QAT in all 12 settings, in 3/3 seeds each, by 1.37 points on average and by 0.20 to 3.50 points; in every setting, the 95\% confidence interval of the paired difference is above zero (\ref{app:stats}). The margin is largest on CIFAR-100: 1.66 to 1.78 points on ResNet-32 and 2.36 to 3.50 on VGG-13. \ours{} is also above PTQ in all 12 settings, by 1.46 to 3.66 points at W8A8. At W2A8, PTQ falls to 8.96--75.23\%, while \ours{} reaches 74.47--93.88\%.

\paragraph{Distance from the teacher}
At W8A8 and W4A8, \ours{} is between 0.06 points below and 0.63 points above its teacher; at W2A8, it is 0.11 to 2.00 points below. With 8-bit activations, the 2-bit students are more accurate than in the benchmark scheme, where the activations also have 2 bits: ResNet-32, for example, reaches 74.47\% at W2A8 and 71.25\% at W2A2 (Tables~\ref{tab:cls-device} and~\ref{tab:cifar}).

\begin{table*}[tbp]
\centering\scriptsize\setlength{\tabcolsep}{1.8pt}
\caption[Classification in the device scheme]{\textbf{CIFAR, device scheme.} Top-1 accuracy (\%) of PTQ, QAT and \ours{} with the final teachers and the first and last layers at 8 bits, mean and standard deviation over three seeds. Gain over QAT: mean paired difference (\ref{app:stats}). Bold: best student in each column.}
\label{tab:cls-device}
\begin{tabular}{@{}lcccccc@{}}
\toprule
\multicolumn{7}{@{}l}{\textbf{(a) CIFAR-10}}\\
 & \multicolumn{3}{c}{ResNet-20} & \multicolumn{3}{c}{VGG-8} \\
 & \multicolumn{3}{c}{\scriptsize FP32 teacher: 94.47} & \multicolumn{3}{c}{\scriptsize FP32 teacher: 93.33} \\
\cmidrule(lr){2-4}\cmidrule(lr){5-7}
Method & W8A8 & W4A8 & W2A8 & W8A8 & W4A8 & W2A8 \\
\midrule
PTQ & 92.79\pmsd{0.04} & 92.09\pmsd{0.05} & 41.42\pmsd{0.04} & 92.24\pmsd{0.02} & 91.90\pmsd{0.03} & 75.23\pmsd{0.10} \\
QAT & 94.26\pmsd{0.14} & 94.21\pmsd{0.08} & 93.52\pmsd{0.16} & 92.98\pmsd{0.21} & 92.91\pmsd{0.05} & 92.54\pmsd{0.11} \\
\rowcolor{oursrow}\ours{} & \textbf{94.57}\pmsd{0.05} & \textbf{94.41}\pmsd{0.11} & \textbf{93.88}\pmsd{0.11} & \textbf{93.70}\pmsd{0.02} & \textbf{93.64}\pmsd{0.03} & \textbf{93.22}\pmsd{0.14} \\
\quad Gain over QAT & {\scriptsize +0.31} & {\scriptsize +0.20} & {\scriptsize +0.36} & {\scriptsize +0.72} & {\scriptsize +0.73} & {\scriptsize +0.69} \\
\midrule[\heavyrulewidth]
\multicolumn{7}{@{}l}{\textbf{(b) CIFAR-100}}\\
 & \multicolumn{3}{c}{ResNet-32} & \multicolumn{3}{c}{VGG-13} \\
 & \multicolumn{3}{c}{\scriptsize FP32 teacher: 76.47} & \multicolumn{3}{c}{\scriptsize FP32 teacher: 78.62} \\
\cmidrule(lr){2-4}\cmidrule(lr){5-7}
Method & W8A8 & W4A8 & W2A8 & W8A8 & W4A8 & W2A8 \\
\midrule
PTQ & 73.34\pmsd{0.02} & 71.66\pmsd{0.04} & 8.96\pmsd{0.12} & 76.09\pmsd{0.01} & 76.12\pmsd{0.02} & 54.76\pmsd{0.07} \\
QAT & 75.30\pmsd{0.18} & 74.77\pmsd{0.30} & 72.69\pmsd{0.41} & 76.86\pmsd{0.04} & 76.75\pmsd{0.27} & 74.87\pmsd{0.14} \\
\rowcolor{oursrow}\ours{} & \textbf{77.00}\pmsd{0.07} & \textbf{76.43}\pmsd{0.09} & \textbf{74.47}\pmsd{0.43} & \textbf{79.25}\pmsd{0.13} & \textbf{79.11}\pmsd{0.09} & \textbf{78.37}\pmsd{0.16} \\
\quad Gain over QAT & {\scriptsize +1.69} & {\scriptsize +1.66} & {\scriptsize +1.78} & {\scriptsize +2.40} & {\scriptsize +2.36} & {\scriptsize +3.50} \\
\bottomrule
\end{tabular}
\end{table*}

\subsection{Segmentation and depth estimation}\label{sec:device-dense}
\paragraph{Comparison with QAT}
On average over three seeds, \ours{} is better than QAT in 15 of the 18 dense settings of Table~\ref{tab:dense-device} ($p = 0.0075$, two-sided sign test over the settings); it is better in 3/3 seeds in 12 of them and in 43 of the 54 paired runs ($p = 1.4\times10^{-5}$, over the runs). The 95\% confidence interval of the paired difference is above zero in 5 of the 18 settings and below zero in none (\ref{app:stats}). The gain reaches 1.16 mIoU on CamVid (R18-PSPNet, W2A8) and 0.0048~m of RMSE on NYUv2 (UNet-small, W8A8). QAT has the better mean in the other three settings, UNet-small on CamVid at W8A8 and W2A8 and Fast-SCNN on NYUv2 at W2A8, with confidence intervals that include zero.

\paragraph{Comparison with PTQ}
\ours{} is better than PTQ in all 18 settings, and the gap is largest at W2A8 for every network and dataset. At W2A8, PTQ falls to 0.41--10.18 mIoU on CamVid and to an RMSE of 1.258--2.383~m on NYUv2, while \ours{} keeps 46.84--65.83 mIoU and an RMSE of 0.592--1.079~m. At W8A8 and W4A8, PTQ comes close to the teacher only with R18-PSPNet (65.33 and 65.02 mIoU against 66.63 on CamVid, 0.595 and 0.602~m against 0.582 on NYUv2); on CamVid, it stays at or below 17.71 mIoU with Fast-SCNN and UNet-small at every bit-width, and with UNet-small at W8A8 it keeps 7.13 mIoU, 15\% of the teacher's 48.12.

\paragraph{Distance from the teacher}
At W8A8 and W4A8, the student of \ours{} exceeds its full-precision teacher in all 12 settings, by 0.14 to 0.78 mIoU and by 0.0058 to 0.0142~m; QAT also exceeds the teacher in 8 of these 12 settings. At W2A8, the students of \ours{} keep on average 95.6\% of the teacher's mIoU on CamVid, and at least 89\% (Fast-SCNN, 5.59 mIoU below the teacher); on NYUv2, their RMSE is lower than that of the teacher with Fast-SCNN and UNet-small and 0.0095~m higher with R18-PSPNet.

\begin{figure*}[tbp]
\centering
\includegraphics[width=\linewidth]{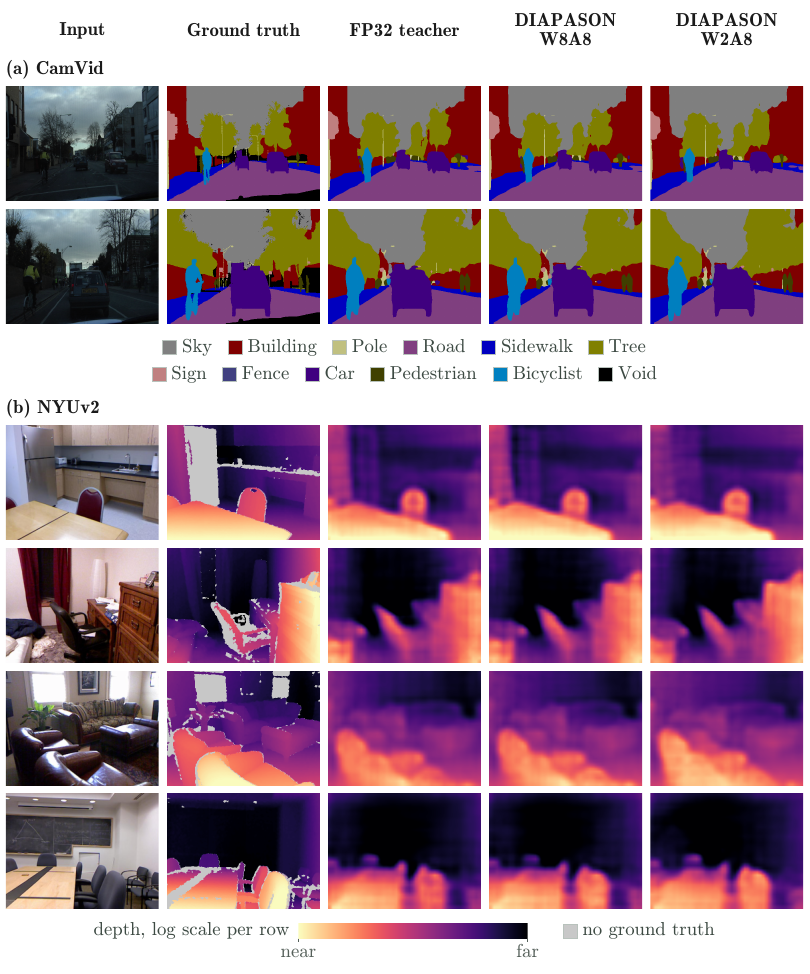}
\caption[Qualitative results]{\textbf{Qualitative results.} Predictions of the FP32 R18-PSPNet teacher and of its \ours{} students at W8A8 and W2A8 in the device scheme (seed~1), on CamVid (a) and NYUv2 (b) test images. Depth on a logarithmic scale per row, from the 1st to the 99th percentile of the ground truth.}
\label{fig:qual}
\end{figure*}

\paragraph{Qualitative results}
Figure~\ref{fig:qual} shows the predictions of the R18-PSPNet students of Table~\ref{tab:dense-device} (seed~1) at W8A8 and W2A8 next to those of the FP32 teacher, on two CamVid and four NYUv2 test images selected for legibility. The students follow the teacher in what it predicts well and in its errors. On CamVid, both students keep the regions predicted by the teacher and differ from it mainly in thin structures such as poles and signs. On NYUv2, the W8A8 maps are close to those of the teacher, and the W2A8 maps keep the layout of each scene with smoother transitions, as around the chairs of the last image; like the teacher, both students give maps smoother than the ground truth and do not resolve thin objects such as chair legs.

\begin{table*}[tbp]
\centering\scriptsize\setlength{\tabcolsep}{1.2pt}
\caption[Segmentation and depth estimation in the device scheme]{\textbf{Segmentation and depth estimation, device scheme.} mIoU on CamVid and RMSE on NYUv2 of PTQ, QAT and \ours{} with the first and last layers at 8 bits, mean and standard deviation over three seeds. Gain over QAT: mean paired difference, positive when \ours{} is better (\ref{app:stats}). Bold: best student in each column.}
\label{tab:dense-device}
\begin{tabular}{@{}lccccccccc@{}}
\toprule
\multicolumn{10}{@{}l}{\textbf{(a) CamVid, mIoU (\%, higher is better)}}\\
 & \multicolumn{3}{c}{Fast-SCNN} & \multicolumn{3}{c}{R18-PSPNet} & \multicolumn{3}{c}{UNet-small} \\
 & \multicolumn{3}{c}{\scriptsize FP32 teacher: 52.43} & \multicolumn{3}{c}{\scriptsize FP32 teacher: 66.63} & \multicolumn{3}{c}{\scriptsize FP32 teacher: 48.12} \\
\cmidrule(lr){2-4}\cmidrule(lr){5-7}\cmidrule(lr){8-10}
Method & W8A8 & W4A8 & W2A8 & W8A8 & W4A8 & W2A8 & W8A8 & W4A8 & W2A8 \\
\midrule
PTQ & 17.71\pmsd{0.37} & 10.22\pmsd{0.31} & 0.41\pmsd{0.00} & 65.33\pmsd{0.13} & 65.02\pmsd{0.01} & 10.18\pmsd{0.21} & 7.13\pmsd{0.09} & 8.08\pmsd{0.21} & 4.25\pmsd{0.30} \\
QAT & 52.05\pmsd{0.31} & 51.75\pmsd{0.29} & 46.31\pmsd{0.63} & 66.62\pmsd{0.19} & 66.55\pmsd{0.03} & 64.68\pmsd{0.16} & \textbf{49.21}\pmsd{0.06} & 48.86\pmsd{0.05} & \textbf{48.59}\pmsd{0.28} \\
\rowcolor{oursrow}\ours{} & \textbf{52.83}\pmsd{0.23} & \textbf{52.70}\pmsd{0.38} & \textbf{46.84}\pmsd{0.22} & \textbf{66.84}\pmsd{0.04} & \textbf{66.77}\pmsd{0.05} & \textbf{65.83}\pmsd{0.15} & 48.64\pmsd{0.21} & \textbf{48.90}\pmsd{0.24} & 47.49\pmsd{0.75} \\
\quad Gain over QAT & {\scriptsize +0.78} & {\scriptsize +0.95} & {\scriptsize +0.53} & {\scriptsize +0.22} & {\scriptsize +0.22} & {\scriptsize +1.16} & {\scriptsize $-$0.56} & {\scriptsize +0.04} & {\scriptsize $-$1.10} \\
\midrule[\heavyrulewidth]
\multicolumn{10}{@{}l}{\textbf{(b) NYUv2, RMSE (m, lower is better)}}\\
 & \multicolumn{3}{c}{Fast-SCNN} & \multicolumn{3}{c}{R18-PSPNet} & \multicolumn{3}{c}{UNet-small} \\
 & \multicolumn{3}{c}{\scriptsize FP32 teacher: 0.815} & \multicolumn{3}{c}{\scriptsize FP32 teacher: 0.582} & \multicolumn{3}{c}{\scriptsize FP32 teacher: 1.084} \\
\cmidrule(lr){2-4}\cmidrule(lr){5-7}\cmidrule(lr){8-10}
Method & W8A8 & W4A8 & W2A8 & W8A8 & W4A8 & W2A8 & W8A8 & W4A8 & W2A8 \\
\midrule
PTQ & 1.247\pmsd{0.034} & 1.171\pmsd{0.037} & 1.357\pmsd{0.034} & 0.595\pmsd{0.002} & 0.602\pmsd{0.002} & 1.258\pmsd{0.007} & 1.848\pmsd{0.018} & 1.850\pmsd{0.003} & 2.383\pmsd{0.022} \\
QAT & 0.803\pmsd{0.002} & 0.806\pmsd{0.001} & \textbf{0.801}\pmsd{0.003} & 0.576\pmsd{0.001} & 0.577\pmsd{0.002} & 0.595\pmsd{0.002} & 1.083\pmsd{0.001} & 1.082\pmsd{0.000} & 1.081\pmsd{0.001} \\
\rowcolor{oursrow}\ours{} & \textbf{0.801}\pmsd{0.001} & \textbf{0.802}\pmsd{0.003} & 0.805\pmsd{0.003} & \textbf{0.574}\pmsd{0.000} & \textbf{0.575}\pmsd{0.002} & \textbf{0.592}\pmsd{0.002} & \textbf{1.078}\pmsd{0.000} & \textbf{1.078}\pmsd{0.001} & \textbf{1.079}\pmsd{0.001} \\
\quad Gain over QAT & {\scriptsize +0.003} & {\scriptsize +0.004} & {\scriptsize $-$0.004} & {\scriptsize +0.001} & {\scriptsize +0.002} & {\scriptsize +0.003} & {\scriptsize +0.005} & {\scriptsize +0.004} & {\scriptsize +0.003} \\
\bottomrule
\end{tabular}
\end{table*}

\section{Deployment on GAP9}\label{sec:gap9}

\paragraph{Deployed models}
The \ours{} students of every setting of Tables~\ref{tab:cls-device} and~\ref{tab:dense-device} are deployed on GAP9: the 12 classification settings with three seeds each and the 18 dense settings with seed~1, 54 models in total. All graphs follow the pipeline of Section~\ref{sec:method-export} and pass both its checks: in ONNX Runtime, they differ from the trained students by $-1.64$ to $+1.23$ points of top-1 accuracy (mean $-0.10$) and by $-0.94$ to $+0.18$ mIoU, and their RMSE is 0.06\% to 1.87\% higher.

\paragraph{Measurement setup}
The 30 settings of Table~\ref{tab:gap9-chip} run on the GAP9 evaluation kit, with seed~1 for the classifiers. The application raises a digital output during each inference; the latency is the duration of this pulse, and the energy is the integral of the measured power over the same interval. Both are averaged over 10 consecutive inferences.

\paragraph{Speed and energy}
Lowering the weights from 8 to 2 bits makes every network faster, by 1.13 to 2.82 times, but the gain depends on the network. The VGG networks gain the most: VGG-8 and VGG-13 run 2.62 and 2.82 times faster at W2A8 than at W8A8, and ResNet-20 and ResNet-32 1.39 and 1.44 times faster. Among the dense models, the speed-up goes from 1.13 times, for Fast-SCNN on both tasks, to 1.58 times, for R18-PSPNet on NYUv2. The energy per inference falls with the latency, and its reduction from W8A8 to W2A8 follows the speed-up: 61\% for VGG-13 and 13\% for Fast-SCNN. In absolute terms, the four CIFAR networks run at 35.8 to 675.7 frames per second; among the dense models, only Fast-SCNN exceeds 30 frames per second, on CamVid at $192\times384$.

\begin{table*}[tbp]
\centering\scriptsize\setlength{\tabcolsep}{4pt}
\caption[Latency and energy on the GAP9 chip]{\textbf{Latency and energy on GAP9.} Latency and energy per inference of the deployed \ours{} students on the GAP9 chip, for one image, mean over 10 consecutive inferences (seed~1). Input: resolution of the deployed graph.}
\label{tab:gap9-chip}
\begin{tabular}{@{}llrrrrrr@{}}
\toprule
 & & \multicolumn{3}{c}{Latency (ms)} & \multicolumn{3}{c}{Energy (mJ)} \\
\cmidrule(lr){3-5}\cmidrule(lr){6-8}
Dataset, network & Input & W8A8 & W4A8 & W2A8 & W8A8 & W4A8 & W2A8 \\
\midrule
CIFAR-10, ResNet-20 & $32\times32$ & 2.06 & 1.70 & 1.48 & 0.134 & 0.108 & 0.089 \\
CIFAR-10, VGG-8 & $32\times32$ & 11.70 & 4.69 & 4.47 & 0.709 & 0.387 & 0.286 \\
CIFAR-100, ResNet-32 & $32\times32$ & 3.37 & 2.76 & 2.34 & 0.222 & 0.176 & 0.143 \\
CIFAR-100, VGG-13 & $32\times32$ & 27.95 & 16.80 & 9.90 & 1.855 & 1.183 & 0.720 \\
\midrule
CamVid, Fast-SCNN & $192\times384$ & 30.58 & 28.48 & 27.01 & 2.060 & 1.925 & 1.801 \\
CamVid, R18-PSPNet & $240\times480$ & 1624.65 & 1314.60 & 1215.72 & 108.935 & 83.316 & 70.508 \\
CamVid, UNet-small & $128\times256$ & 238.82 & 209.69 & 181.17 & 16.504 & 13.893 & 11.162 \\
\midrule
NYUv2, Fast-SCNN & $288\times384$ & 39.18 & 36.74 & 34.76 & 2.691 & 2.519 & 2.350 \\
NYUv2, R18-PSPNet & $288\times384$ & 1564.11 & 1100.34 & 992.52 & 102.594 & 71.016 & 59.776 \\
NYUv2, UNet-small & $288\times384$ & 953.84 & 781.40 & 737.74 & 61.724 & 49.992 & 43.397 \\
\bottomrule
\end{tabular}
\end{table*}

\begin{figure*}[tbp]
\centering
\includegraphics[width=\linewidth]{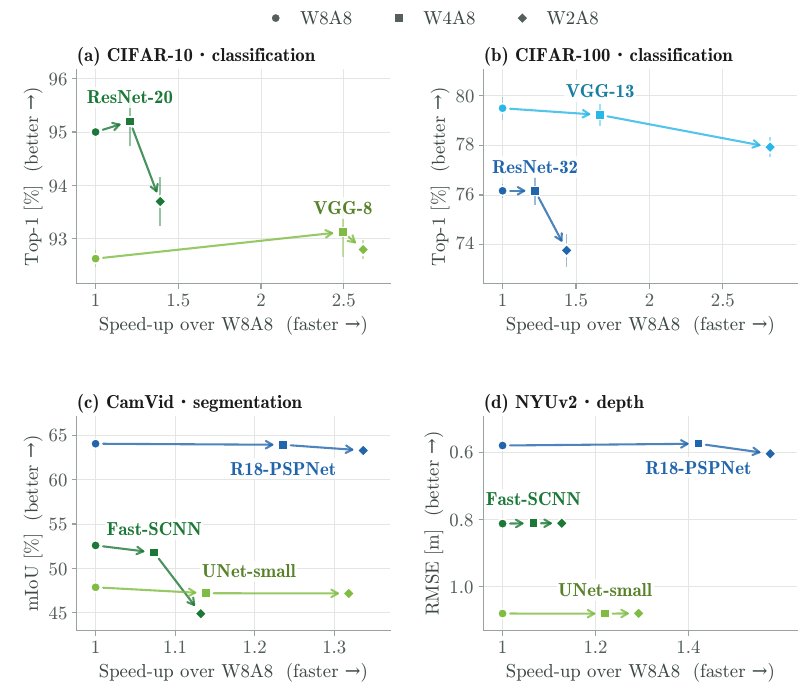}
\caption[Task metric versus speed-up on GAP9]{\textbf{Task metric and speed-up on GAP9.} Task metric of the exported \ours{} students against their speed-up over W8A8 on the GAP9 chip. Task metric from GVSOC for the classifiers (1000 test images, three seeds) and from ONNX Runtime for the dense models (full test set, seed~1).}
\label{fig:pareto}
\end{figure*}

\paragraph{Task metric of the exported graphs}
Figure~\ref{fig:pareto} relates the speed-up over W8A8 on the chip to the task metric of the exported graph. For the classifiers, this metric comes from GVSOC on 1000 test images, as mean and standard deviation over three seeds; for the dense models, it comes from ONNX Runtime on the full test set, at the input resolution of Table~\ref{tab:gap9-chip}, with seed~1. On CamVid, this resolution is lower than the test resolution of Table~\ref{tab:dense-device}, so the mIoU in Figure~\ref{fig:pareto} is not directly comparable with that table.
\paragraph{From 8 to 4 and from 4 to 2 bits}
The step from W8A8 to W4A8 costs little accuracy: every network becomes 1.07 to 2.49 times faster, while the top-1 accuracy or mIoU decreases by at most 0.81 points and the RMSE increases by at most 0.0004~m. The step from W4A8 to W2A8 gives a further speed-up of 1.05 to 1.70 times, at a cost of up to 2.40 points of top-1 accuracy (ResNet-32), 6.89 mIoU (Fast-SCNN on CamVid) and 0.030~m of RMSE (R18-PSPNet on NYUv2). Whether this second step pays off depends on the network: UNet-small on CamVid becomes 1.16 times faster for 0.02 mIoU, and VGG-13 1.70 times faster for 1.30 points of top-1 accuracy, whereas Fast-SCNN on CamVid gains only 1.05 times for the largest loss in mIoU.

\section{Discussion, Limitations, and Future Directions}\label{sec:discussion}
\paragraph{Two-bit weights}
With 8-bit activations, the W2A8 students of \ours{} keep on average 97.6\% of their teacher's task metric in classification and segmentation, and at least 89\% (Tables~\ref{tab:cls-device} and~\ref{tab:dense-device}); on GAP9, 2-bit weights make the same networks 1.13 to 2.82 times faster than 8-bit weights (Section~\ref{sec:gap9}).

\paragraph{Effect of the teacher}
The comparison with published results mixes two effects: that of the method, including its training setup (TrivialAugment, the initialization of the ranges from percentiles and the straight-through gradient), and that of the teacher (Section~\ref{sec:bench-cifar}). The \reft{} teachers reduce the second effect without removing it: in the \reft{}-teacher column of Table~\ref{tab:cifar}, the teacher accuracy is within 0.91 points of that of SQAKD and SQuaT. The FP32 rows also show the teachers of DA-QAKD next to ours.

\paragraph{Labels in the classification objective}
The classification objective of Eq.~\eqref{eq:cls} uses labels, so the classification results do not measure distillation without labels, which is the setting of SQAKD. \ref{app:ablations} compares Eq.~\eqref{eq:cls} with each of its two terms alone, under the same teacher; there, with the TrivialAugment teachers, the cross-entropy alone has the highest mean in 5 of the 12 CIFAR settings of the benchmark scheme, all at W8A8 or W4A4, whereas at W2A2 the combined objective is above the cross-entropy alone in all four networks, by 0.22 to 1.74 points.

\paragraph{Selection of the training setup}
Four choices come from single-seed runs evaluated on the test sets: the combination of cross-entropy, distillation and TrivialAugment, the one-batch initialization at 1 bit, the setup of DeiT-Tiny at W2A2 and the bit-dependent learning rate of the dense students.

\paragraph{Dense tasks}
On the dense tasks, \ours{} is better than QAT in 15 of the 18 settings ($p = 0.0075$) and in 43 of the 54 paired runs, but with three seeds the confidence interval of the paired difference is above zero in 5 of the 18 settings and includes zero in the other 13 (\ref{app:stats}). In the leave-one-out at W2A2 in the benchmark scheme, the holistic term contributes most: removing it lowers the mIoU by 1.55 and 5.12 points in the two segmentation networks tested, against 0.11 and 0.85 points for the pair-wise term (\ref{app:ablations}). The dense students exceed their teachers at W8A8 and W4A8; this suggests that the comparison with the teacher measures the additional training from the teacher weights as well as the cost of quantization.

\paragraph{Tiny-ImageNet}
On Tiny-ImageNet, the teachers start from ImageNet-1k weights and Tiny-ImageNet is drawn from ImageNet \cite{le2015tinyimagenet}, so the pretraining data may overlap with the validation images. The QAT baseline is below PTQ in 5 of these 9 settings (Section~\ref{sec:bench-tiny}), so the margins of \ours{} over it may overstate the effect of distillation. Tiny-ImageNet is evaluated in the benchmark scheme only, so none of its models is deployed.

\paragraph{Measurements of the deployed models}
The task metrics of the deployed models come from ONNX Runtime and GVSOC, not from the chip; on the chip, only the latency and the energy are measured, for seed~1 and over 10 inferences per setting. The dense models are deployed for one seed only. Activations below 8 bits and binary weights are outside the device scheme.

\paragraph{Future work}
Future work will bring below 8 bits the activations and the input and output of the network, which the device scheme keeps at 8 bits, and will port the deployment pipeline to other microcontrollers with neural accelerators.

\section{Conclusion}\label{sec:conclusion}

\ours{} trains quantized networks for classification, semantic segmentation and depth estimation by quantization-aware self-distillation, with pixel-wise and pair-wise terms in the dense tasks and a holistic term in segmentation. On CIFAR-10 and CIFAR-100, it improves the best published result in 9 of 12 configurations. Beyond this benchmark, it is more accurate than the QAT baseline in all 21 classification settings; in segmentation and depth estimation, it has a better mean in 15 of the 18 settings and is better in 43 of the 54 paired runs. The students run on GAP9 as integer networks: 54 students are deployed, and the latency and energy of all 30 settings are measured on the chip, where 2-bit weights speed up inference up to 2.82 times over 8-bit weights.

The main limitations are the dense tasks, where the confidence interval of the difference with QAT is above zero in 5 of the 18 settings, the use of the test sets to select the training setup, and the task metric of the deployed models, which is measured in GVSOC and ONNX Runtime, not on the chip.

\section*{Acknowledgements}
This work was carried out in the EssilorLuxottica Smart Eyewear Lab, a Joint Research Center between EssilorLuxottica and Politecnico di Milano.

\bibliographystyle{plainnat}
\bibliography{references}

\begin{thebibliography}{68}
\providecommand{\natexlab}[1]{#1}
\providecommand{\url}[1]{\texttt{#1}}
\expandafter\ifx\csname urlstyle\endcsname\relax
  \providecommand{\doi}[1]{doi: #1}\else
  \providecommand{\doi}{doi: \begingroup \urlstyle{rm}\Url}\fi

\bibitem[{Analog Devices}(2024)]{adimax78000}
{Analog Devices}.
\newblock {ADI} {MAX78000/MAX78002} model training and synthesis.
\newblock \url{https://github.com/analogdevicesinc/ai8x-synthesis}, 2024.
\newblock Accessed 21 November 2026.

\bibitem[{Arm}(2026)]{armvela}
{Arm}.
\newblock {Vela}: Supported ops, {Vela} 5.2.0.
\newblock
  \url{https://gitlab.arm.com/artificial-intelligence/ethos-u/ethos-u-vela/-/raw/5.2.0/SUPPORTED_OPS.md},
  2026.
\newblock Accessed 21 November 2026.

\bibitem[Banbury et~al.(2021)Banbury, Janapa~Reddi, Torelli, Holleman,
  Jeffries, Kiraly, Montino, Kanter, Ahmed, Pau, Thakker, Torrini, Warden,
  Cordaro, Di~Guglielmo, Duarte, Gibellini, Parekh, Tran, Tran, Niu, and
  Xu]{banbury2021mlperftiny}
Colby Banbury, Vijay Janapa~Reddi, Peter Torelli, Jeremy Holleman, Nat
  Jeffries, Csaba Kiraly, Pietro Montino, David Kanter, Sebastian Ahmed, Danilo
  Pau, Urmish Thakker, Antonio Torrini, Pete Warden, Jay Cordaro, Giuseppe
  Di~Guglielmo, Javier Duarte, Stephen Gibellini, Videet Parekh, Honson Tran,
  Nhan Tran, Wenxu Niu, and Xuesong Xu.
\newblock {MLPerf} tiny benchmark.
\newblock In \emph{Proceedings of the Neural Information Processing Systems
  Track on Datasets and Benchmarks}, volume~1, 2021.
\newblock arXiv:2106.07597.

\bibitem[Bengio et~al.(2013)Bengio, L{\'e}onard, and Courville]{bengio2013ste}
Yoshua Bengio, Nicholas L{\'e}onard, and Aaron Courville.
\newblock Estimating or propagating gradients through stochastic neurons for
  conditional computation, 2013.

\bibitem[Boo et~al.(2021)Boo, Shin, Choi, and Sung]{boo2021speq}
Yoonho Boo, Sungho Shin, Jungwook Choi, and Wonyong Sung.
\newblock Stochastic precision ensemble: Self-knowledge distillation for
  quantized deep neural networks.
\newblock In \emph{Proceedings of the AAAI Conference on Artificial
  Intelligence}, volume~35, pages 6794--6802, 2021.
\newblock \doi{10.1609/aaai.v35i8.16839}.
\newblock arXiv:2009.14502.

\bibitem[Brostow et~al.(2009)Brostow, Fauqueur, and Cipolla]{brostow2009camvid}
Gabriel~J. Brostow, Julien Fauqueur, and Roberto Cipolla.
\newblock Semantic object classes in video: A high-definition ground truth
  database.
\newblock \emph{Pattern Recognition Letters}, 30\penalty0 (2):\penalty0 88--97,
  2009.
\newblock \doi{10.1016/j.patrec.2008.04.005}.

\bibitem[Bruschi et~al.(2021)Bruschi, Haugou, Tagliavini, Conti, Benini, and
  Rossi]{bruschi2021gvsoc}
Nazareno Bruschi, Germain Haugou, Giuseppe Tagliavini, Francesco Conti, Luca
  Benini, and Davide Rossi.
\newblock {GVSoC}: A highly configurable, fast and accurate full-platform
  simulator for {RISC-V} based {IoT} processors.
\newblock In \emph{Proceedings of the IEEE 39th International Conference on
  Computer Design (ICCD)}, pages 409--416, 2021.
\newblock \doi{10.1109/ICCD53106.2021.00071}.

\bibitem[Cho and Hariharan(2019)]{cho2019efficacy}
Jang~Hyun Cho and Bharath Hariharan.
\newblock On the efficacy of knowledge distillation.
\newblock In \emph{Proceedings of the IEEE/CVF International Conference on
  Computer Vision (ICCV)}, pages 4794--4802, 2019.

\bibitem[Choi et~al.(2026)Choi, Yang, Lim, Jeong, Zinke, Yu, and
  Kang]{choi2026spaq}
Jaemin Choi, Jincheol Yang, Nahyun Lim, Yun-Seong Jeong, Matti Zinke, Hyunwoo
  Yu, and Suk-Ju Kang.
\newblock {SPA-Q}: Structure-preserving adaptive post-training quantization for
  monocular depth estimation.
\newblock In \emph{Advances in Neural Information Processing Systems}, 2026.

\bibitem[Choi et~al.(2018)Choi, Wang, Venkataramani, Chuang, Srinivasan, and
  Gopalakrishnan]{choi2018pact}
Jungwook Choi, Zhuo Wang, Swagath Venkataramani, Pierce I-Jen Chuang,
  Vijayalakshmi Srinivasan, and Kailash Gopalakrishnan.
\newblock {PACT}: Parameterized clipping activation for quantized neural
  networks, 2018.

\bibitem[Cordts et~al.(2016)Cordts, Omran, Ramos, Rehfeld, Enzweiler, Benenson,
  Franke, Roth, and Schiele]{cordts2016cityscapes}
Marius Cordts, Mohamed Omran, Sebastian Ramos, Timo Rehfeld, Markus Enzweiler,
  Rodrigo Benenson, Uwe Franke, Stefan Roth, and Bernt Schiele.
\newblock The {Cityscapes} dataset for semantic urban scene understanding.
\newblock In \emph{Proceedings of the IEEE Conference on Computer Vision and
  Pattern Recognition (CVPR)}, pages 3213--3223, 2016.

\bibitem[Cubuk et~al.(2019)Cubuk, Zoph, Man{\'e}, Vasudevan, and
  Le]{cubuk2019autoaugment}
Ekin~D. Cubuk, Barret Zoph, Dandelion Man{\'e}, Vijay Vasudevan, and Quoc~V.
  Le.
\newblock {AutoAugment}: Learning augmentation strategies from data.
\newblock In \emph{Proceedings of the IEEE/CVF Conference on Computer Vision
  and Pattern Recognition (CVPR)}, pages 113--123, 2019.

\bibitem[Deng et~al.(2009)Deng, Dong, Socher, Li, Li, and
  Fei-Fei]{deng2009imagenet}
Jia Deng, Wei Dong, Richard Socher, Li-Jia Li, Kai Li, and Li~Fei-Fei.
\newblock {ImageNet}: A large-scale hierarchical image database.
\newblock In \emph{Proceedings of the IEEE Conference on Computer Vision and
  Pattern Recognition (CVPR)}, pages 248--255, 2009.
\newblock \doi{10.1109/CVPR.2009.5206848}.

\bibitem[DeVries and Taylor(2017)]{devries2017cutout}
Terrance DeVries and Graham~W. Taylor.
\newblock Improved regularization of convolutional neural networks with cutout,
  2017.

\bibitem[Esser et~al.(2020)Esser, McKinstry, Bablani, Appuswamy, and
  Modha]{esser2020lsq}
Steven~K. Esser, Jeffrey~L. McKinstry, Deepika Bablani, Rathinakumar Appuswamy,
  and Dharmendra~S. Modha.
\newblock Learned step size quantization.
\newblock In \emph{International Conference on Learning Representations
  (ICLR)}, 2020.
\newblock arXiv:1902.08153.

\bibitem[{GreenWaves Technologies}(2022)]{gapsdknntool}
{GreenWaves Technologies}.
\newblock {GAP} {SDK}: {NNTool} documentation, release v4.22.0.
\newblock
  \url{https://github.com/GreenWaves-Technologies/gap_sdk/tree/release-v4.22.0/tools/nntool},
  2022.
\newblock Accessed 21 November 2026.

\bibitem[{GreenWaves Technologies}(2023)]{gapquantimport}
{GreenWaves Technologies}.
\newblock {CIFAR10} on {GAP}.
\newblock
  \url{https://github.com/GreenWaves-Technologies/cifar10_quant_import}, 2023.
\newblock Accessed 21 November 2026.

\bibitem[{GreenWaves Technologies}(2024)]{gap9}
{GreenWaves Technologies}.
\newblock {GAP9} datasheet, version 1.1.
\newblock
  \url{https://web.archive.org/web/20250126112055/https://greenwaves-technologies.com/wp-content/uploads/2024/07/gap9datasheet_v1.1.pdf},
  2024.
\newblock {Internet Archive} copy of 26 January 2025, accessed 21 November
  2026.

\bibitem[Gulrajani et~al.(2017)Gulrajani, Ahmed, Arjovsky, Dumoulin, and
  Courville]{gulrajani2017wgangp}
Ishaan Gulrajani, Faruk Ahmed, Martin Arjovsky, Vincent Dumoulin, and Aaron
  Courville.
\newblock Improved training of {W}asserstein {GAN}s.
\newblock In \emph{Advances in Neural Information Processing Systems},
  volume~30. Curran Associates, Inc., 2017.

\bibitem[He et~al.(2016)He, Zhang, Ren, and Sun]{he2016resnet}
Kaiming He, Xiangyu Zhang, Shaoqing Ren, and Jian Sun.
\newblock Deep residual learning for image recognition.
\newblock In \emph{Proceedings of the IEEE Conference on Computer Vision and
  Pattern Recognition (CVPR)}, pages 770--778, 2016.

\bibitem[He et~al.(2026)He, Guo, Li, Cui, Weng, Li, and
  Zhang]{he2026distillanydepth}
Xiankang He, Dongyan Guo, Hongji Li, Ying Cui, Libo Weng, Ruibo Li, and Chi
  Zhang.
\newblock Distill any depth: Distillation creates a stronger monocular depth
  estimator.
\newblock In \emph{Proceedings of the IEEE/CVF Conference on Computer Vision
  and Pattern Recognition (CVPR) Findings}, pages 591--601, 2026.
\newblock \doi{10.1109/CVPRF71909.2026.00061}.

\bibitem[Hinton et~al.(2015)Hinton, Vinyals, and Dean]{hinton2015kd}
Geoffrey Hinton, Oriol Vinyals, and Jeff Dean.
\newblock Distilling the knowledge in a neural network, 2015.
\newblock {NIPS} 2014 Deep Learning Workshop.

\bibitem[Jacob et~al.(2018)Jacob, Kligys, Chen, Zhu, Tang, Howard, Adam, and
  Kalenichenko]{jacob2018integer}
Benoit Jacob, Skirmantas Kligys, Bo~Chen, Menglong Zhu, Matthew Tang, Andrew
  Howard, Hartwig Adam, and Dmitry Kalenichenko.
\newblock Quantization and training of neural networks for efficient
  integer-arithmetic-only inference.
\newblock In \emph{Proceedings of the IEEE Conference on Computer Vision and
  Pattern Recognition (CVPR)}, pages 2704--2713, 2018.

\bibitem[Khan et~al.(2025)Khan, Iftikhar, Ward, Davy, and
  Breslin]{khan2025biagrinet}
Hassan Khan, Sunbal Iftikhar, Rory Ward, Steven Davy, and John~G. Breslin.
\newblock {BiAgriNet}: Binarized knowledge-distilled network for real-time
  semantic segmentation in agriculture.
\newblock \emph{IEEE Access}, 13:\penalty0 192374--192390, 2025.
\newblock \doi{10.1109/ACCESS.2025.3629617}.

\bibitem[Kim et~al.(2019)Kim, Bhalgat, Lee, Patel, and Kwak]{kim2019qkd}
Jangho Kim, Yash Bhalgat, Jinwon Lee, Chirag Patel, and Nojun Kwak.
\newblock {QKD}: Quantization-aware knowledge distillation, 2019.

\bibitem[Krizhevsky(2009)]{krizhevsky2009cifar}
Alex Krizhevsky.
\newblock Learning multiple layers of features from tiny images.
\newblock Technical report, University of Toronto, 2009.
\newblock URL
  \url{https://www.cs.toronto.edu/~kriz/learning-features-2009-TR.pdf}.

\bibitem[Kur et~al.(2026)Kur, Maravilla, and Zhao]{kur2025daqakd}
Justin Kur, Jon Maravilla, and Kaiqi Zhao.
\newblock Data-augmented quantization-aware knowledge distillation.
\newblock In \emph{Proceedings of the International Joint Conference on Neural
  Networks (IJCNN)}, pages 1--8, 2026.
\newblock \doi{10.1109/IJCNN69854.2026.11729846}.

\bibitem[Laina et~al.(2016)Laina, Rupprecht, Belagiannis, Tombari, and
  Navab]{laina2016berhu}
Iro Laina, Christian Rupprecht, Vasileios Belagiannis, Federico Tombari, and
  Nassir Navab.
\newblock Deeper depth prediction with fully convolutional residual networks.
\newblock In \emph{Proceedings of the Fourth International Conference on 3D
  Vision (3DV)}, pages 239--248, 2016.
\newblock \doi{10.1109/3DV.2016.32}.

\bibitem[Le and Yang(2015)]{le2015tinyimagenet}
Ya~Le and Xuan Yang.
\newblock Tiny {ImageNet} visual recognition challenge, 2015.
\newblock URL
  \url{https://cs231n.stanford.edu/reports/2015/pdfs/yle_project.pdf}.
\newblock {CS231n} course report, Stanford University.

\bibitem[Lee et~al.(2026)Lee, Jo, Chung, and Kim]{lee2026squat}
HyeonJun Lee, Hyeonsik Jo, Jinwoo Chung, and Jangho Kim.
\newblock {SQuaT}: Self-supervised knowledge distillation via student-aware
  quantized teacher features.
\newblock In \emph{Proceedings of the 29th International Conference on
  Artificial Intelligence and Statistics (AISTATS)}, volume 300 of
  \emph{Proceedings of Machine Learning Research}, pages 5023--5031, 2026.
\newblock arXiv:2608.10709.

\bibitem[Lee et~al.(2021)Lee, Kim, and Ham]{lee2021ewgs}
Junghyup Lee, Dohyung Kim, and Bumsub Ham.
\newblock Network quantization with element-wise gradient scaling.
\newblock In \emph{Proceedings of the IEEE/CVF Conference on Computer Vision
  and Pattern Recognition (CVPR)}, pages 6448--6457, 2021.

\bibitem[Li et~al.(2021)Li, Shen, Ma, Ren, Zhao, Zhang, Gong, Yu, and
  Yan]{li2021mqbench}
Yuhang Li, Mingzhu Shen, Jian Ma, Yan Ren, Mingxin Zhao, Qi~Zhang, Ruihao Gong,
  Fengwei Yu, and Junjie Yan.
\newblock {MQBench}: Towards reproducible and deployable model quantization
  benchmark.
\newblock In \emph{Proceedings of the Neural Information Processing Systems
  Track on Datasets and Benchmarks}, volume~1, 2021.
\newblock arXiv:2111.03759.

\bibitem[Lin et~al.(2020)Lin, Chen, Lin, Cohn, Gan, and Han]{lin2020mcunet}
Ji~Lin, Wei-Ming Chen, Yujun Lin, John Cohn, Chuang Gan, and Song Han.
\newblock {MCUNet}: Tiny deep learning on {IoT} devices.
\newblock In \emph{Advances in Neural Information Processing Systems},
  volume~33, pages 11711--11722. Curran Associates, Inc., 2020.

\bibitem[Liu et~al.(2019)Liu, Chen, Liu, Qin, Luo, and Wang]{liu2019skd}
Yifan Liu, Ke~Chen, Chris Liu, Zengchang Qin, Zhenbo Luo, and Jingdong Wang.
\newblock Structured knowledge distillation for semantic segmentation.
\newblock In \emph{Proceedings of the IEEE/CVF Conference on Computer Vision
  and Pattern Recognition (CVPR)}, pages 2604--2613, 2019.

\bibitem[Liu et~al.(2023)Liu, Shu, Wang, and Shen]{liu2020skd}
Yifan Liu, Changyong Shu, Jingdong Wang, and Chunhua Shen.
\newblock Structured knowledge distillation for dense prediction.
\newblock \emph{IEEE Transactions on Pattern Analysis and Machine
  Intelligence}, 45\penalty0 (6):\penalty0 7035--7049, 2023.
\newblock \doi{10.1109/TPAMI.2020.3001940}.
\newblock Preprint: arXiv:1903.04197v7.

\bibitem[Liu et~al.(2021)Liu, Zhang, and Wang]{liu2021zaq}
Yuang Liu, Wei Zhang, and Jun Wang.
\newblock Zero-shot adversarial quantization.
\newblock In \emph{Proceedings of the IEEE/CVF Conference on Computer Vision
  and Pattern Recognition (CVPR)}, pages 1512--1521, 2021.
\newblock arXiv:2103.15263.

\bibitem[Lu et~al.(2022)Lu, Zeng, and Zeng]{lu2022quantdepth}
Keyu Lu, Chengyi Zeng, and Yonghu Zeng.
\newblock Self-supervised learning of monocular depth using quantized networks.
\newblock \emph{Neurocomputing}, 488:\penalty0 634--646, 2022.
\newblock \doi{10.1016/j.neucom.2021.11.071}.

\bibitem[Millar et~al.(2025)Millar, Huang, Sethi, Haddadi, and
  Madhavapeddy]{millar2025munpu}
Josh Millar, Yushan Huang, Sarab Sethi, Hamed Haddadi, and Anil Madhavapeddy.
\newblock Benchmarking ultra-low-power {$\mu$NPUs}.
\newblock In \emph{Proceedings of the 31st Annual International Conference on
  Mobile Computing and Networking (MobiCom)}, pages 1060--1074, 2025.
\newblock \doi{10.1145/3680207.3765264}.

\bibitem[Mishra and Marr(2018)]{mishra2018apprentice}
Asit Mishra and Debbie Marr.
\newblock Apprentice: Using knowledge distillation techniques to improve
  low-precision network accuracy.
\newblock In \emph{International Conference on Learning Representations
  (ICLR)}, 2018.

\bibitem[M{\"u}ller and Hutter(2021)]{muller2021trivialaugment}
Samuel~G. M{\"u}ller and Frank Hutter.
\newblock {TrivialAugment}: Tuning-free yet state-of-the-art data augmentation.
\newblock In \emph{Proceedings of the IEEE/CVF International Conference on
  Computer Vision (ICCV)}, pages 774--782, 2021.

\bibitem[Nagel et~al.(2021)Nagel, Fournarakis, Amjad, Bondarenko, van Baalen,
  and Blankevoort]{nagel2021whitepaper}
Markus Nagel, Marios Fournarakis, Rana~Ali Amjad, Yelysei Bondarenko, Mart van
  Baalen, and Tijmen Blankevoort.
\newblock A white paper on neural network quantization, 2021.

\bibitem[{ONNX Runtime developers}(2026)]{onnxruntime}
{ONNX Runtime developers}.
\newblock {ONNX Runtime}.
\newblock \url{https://onnxruntime.ai}, 2026.
\newblock Version 1.27.0, accessed 21 November 2026.

\bibitem[Pham et~al.(2023)Pham, Hoang, and Do]{pham2023cmtkd}
Cuong Pham, Tuan Hoang, and Thanh-Toan Do.
\newblock Collaborative multi-teacher knowledge distillation for learning low
  bit-width deep neural networks.
\newblock In \emph{Proceedings of the IEEE/CVF Winter Conference on
  Applications of Computer Vision (WACV)}, pages 6435--6443, 2023.
\newblock arXiv:2210.16103.

\bibitem[Pizzolato et~al.(2026)Pizzolato, Pasti, and
  Bellotto]{pizzolato2026nanou}
Federico Pizzolato, Francesco Pasti, and Nicola Bellotto.
\newblock {Nano-U}: Efficient terrain segmentation for tiny robot navigation,
  2026.
\newblock Accepted at TAROS 2026.

\bibitem[Poggi et~al.(2020)Poggi, Aleotti, Tosi, and
  Mattoccia]{poggi2020uncertainty}
Matteo Poggi, Filippo Aleotti, Fabio Tosi, and Stefano Mattoccia.
\newblock On the uncertainty of self-supervised monocular depth estimation.
\newblock In \emph{Proceedings of the IEEE/CVF Conference on Computer Vision
  and Pattern Recognition (CVPR)}, pages 3227--3237, 2020.

\bibitem[Poudel et~al.(2019)Poudel, Liwicki, and Cipolla]{poudel2019fastscnn}
Rudra P.~K. Poudel, Stephan Liwicki, and Roberto Cipolla.
\newblock {Fast-SCNN}: Fast semantic segmentation network.
\newblock In \emph{Proceedings of the British Machine Vision Conference
  (BMVC)}, 2019.
\newblock arXiv:1902.04502.

\bibitem[{PULP Platform}(2022)]{ne16repo}
{PULP Platform}.
\newblock Neural engine 16-channels.
\newblock \url{https://github.com/pulp-platform/ne16}, 2022.
\newblock Accessed 21 November 2026.

\bibitem[Ronneberger et~al.(2015)Ronneberger, Fischer, and
  Brox]{ronneberger2015unet}
Olaf Ronneberger, Philipp Fischer, and Thomas Brox.
\newblock {U-Net}: Convolutional networks for biomedical image segmentation.
\newblock In \emph{Medical Image Computing and Computer-Assisted Intervention
  -- MICCAI 2015}, volume 9351 of \emph{Lecture Notes in Computer Science},
  pages 234--241. Springer, 2015.
\newblock \doi{10.1007/978-3-319-24574-4_28}.

\bibitem[Rusci et~al.(2020)Rusci, Capotondi, and Benini]{rusci2020mixed}
Manuele Rusci, Alessandro Capotondi, and Luca Benini.
\newblock Memory-driven mixed low precision quantization for enabling deep
  network inference on microcontrollers.
\newblock In \emph{Proceedings of Machine Learning and Systems (MLSys)},
  volume~2, pages 326--335, 2020.
\newblock arXiv:1905.13082.

\bibitem[Saha et~al.(2022)Saha, Sandha, and Srivastava]{saha2022mcureview}
Swapnil~Sayan Saha, Sandeep~Singh Sandha, and Mani Srivastava.
\newblock Machine learning for microcontroller-class hardware: A review.
\newblock \emph{IEEE Sensors Journal}, 22\penalty0 (22):\penalty0 21362--21390,
  2022.
\newblock \doi{10.1109/JSEN.2022.3210773}.
\newblock arXiv:2205.14550.

\bibitem[Sandler et~al.(2018)Sandler, Howard, Zhu, Zhmoginov, and
  Chen]{sandler2018mobilenetv2}
Mark Sandler, Andrew Howard, Menglong Zhu, Andrey Zhmoginov, and Liang-Chieh
  Chen.
\newblock {MobileNetV2}: Inverted residuals and linear bottlenecks.
\newblock In \emph{Proceedings of the IEEE Conference on Computer Vision and
  Pattern Recognition (CVPR)}, pages 4510--4520, 2018.

\bibitem[Shen et~al.(2025)Shen, Ma, Liu, Yang, Ding, Wang, Ding, Niu, Wang,
  Zhao, Lin, and Gu]{shen2025quartdepth}
Xuan Shen, Weize Ma, Jing Liu, Changdi Yang, Rui Ding, Quanyi Wang, Henghui
  Ding, Wei Niu, Yanzhi Wang, Pu~Zhao, Jun Lin, and Jiuxiang Gu.
\newblock {QuartDepth}: Post-training quantization for real-time depth
  estimation on the edge.
\newblock In \emph{Proceedings of the IEEE/CVF Conference on Computer Vision
  and Pattern Recognition (CVPR)}, pages 11448--11460, 2025.
\newblock arXiv:2503.16709.

\bibitem[Shu et~al.(2021)Shu, Liu, Gao, Yan, and Shen]{shu2021cwd}
Changyong Shu, Yifan Liu, Jianfei Gao, Zheng Yan, and Chunhua Shen.
\newblock Channel-wise knowledge distillation for dense prediction.
\newblock In \emph{Proceedings of the IEEE/CVF International Conference on
  Computer Vision (ICCV)}, pages 5311--5320, 2021.
\newblock arXiv:2011.13256.

\bibitem[Siddappa et~al.(2026)Siddappa, Eskandari, Dasan, Sharma, and
  Thonse]{tinpuguide}
P.~Siddappa, N.~Eskandari, N.~Dasan, T.~Sharma, and A.~Thonse.
\newblock Neural-network processing unit ({NPU}) guide.
\newblock Application Note SDAA185, Texas Instruments, 2026.
\newblock URL \url{https://www.ti.com/lit/an/sdaa185/sdaa185.pdf}.
\newblock Accessed 21 November 2026.

\bibitem[Silberman et~al.(2012)Silberman, Hoiem, Kohli, and
  Fergus]{silberman2012nyu}
Nathan Silberman, Derek Hoiem, Pushmeet Kohli, and Rob Fergus.
\newblock Indoor segmentation and support inference from {RGBD} images.
\newblock In \emph{Computer Vision -- ECCV 2012}, volume 7576 of \emph{Lecture
  Notes in Computer Science}, pages 746--760. Springer, 2012.
\newblock \doi{10.1007/978-3-642-33715-4_54}.

\bibitem[Simonyan and Zisserman(2015)]{simonyan2015vgg}
Karen Simonyan and Andrew Zisserman.
\newblock Very deep convolutional networks for large-scale image recognition.
\newblock In \emph{International Conference on Learning Representations
  (ICLR)}, 2015.

\bibitem[{STMicroelectronics}(2026)]{stneuralart}
{STMicroelectronics}.
\newblock {Neural-ART}: Supported operators and limitations, r1.4, {ST Edge AI
  Core} 4.1.0.
\newblock
  \url{https://stedgeai-dc.st.com/assets/embedded-docs/stneuralart_operator_support.html},
  2026.
\newblock Accessed 21 November 2026.

\bibitem[Touvron et~al.(2021)Touvron, Cord, Douze, Massa, Sablayrolles, and
  J{\'e}gou]{touvron2021deit}
Hugo Touvron, Matthieu Cord, Matthijs Douze, Francisco Massa, Alexandre
  Sablayrolles, and Herv{\'e} J{\'e}gou.
\newblock Training data-efficient image transformers \& distillation through
  attention.
\newblock In \emph{Proceedings of the 38th International Conference on Machine
  Learning (ICML)}, volume 139 of \emph{Proceedings of Machine Learning
  Research}, pages 10347--10357, 2021.

\bibitem[Wang and Wang(2026)]{wang2026q2}
Zhaoyang Wang and Dong Wang.
\newblock {Q$^2$}: Quantization-aware gradient balancing and attention
  alignment for low-bit quantization, 2026.
\newblock Preprint, version 2.

\bibitem[Yang et~al.(2025)Yang, Choi, Zinke, and Kang]{yang2025qsca}
Jincheol Yang, Jaemin Choi, Matti Zinke, and Suk-Ju Kang.
\newblock {QSCA}: Quantization with self-compensating auxiliary for monocular
  depth estimation.
\newblock In \emph{Advances in Neural Information Processing Systems},
  volume~38, pages 31444--31465. Curran Associates, Inc., 2025.
\newblock \doi{10.52202/085713-1055}.

\bibitem[Yu et~al.(2023)Yu, Chen, Gan, and Fan]{yu2023gpusqvit}
Chong Yu, Tao Chen, Zhongxue Gan, and Jiayuan Fan.
\newblock Boost vision transformer with {GPU}-friendly sparsity and
  quantization.
\newblock In \emph{Proceedings of the IEEE/CVF Conference on Computer Vision
  and Pattern Recognition (CVPR)}, pages 22658--22668, 2023.
\newblock \doi{10.1109/CVPR52729.2023.02170}.

\bibitem[Yu et~al.(2020)Yu, Wen, Cheng, Sun, Han, and Shi]{yu2019gdrq}
Haibao Yu, Tuopu Wen, Guangliang Cheng, Jiankai Sun, Qi~Han, and Jianping Shi.
\newblock Low-bit quantization needs good distribution.
\newblock In \emph{Proceedings of the IEEE/CVF Conference on Computer Vision
  and Pattern Recognition (CVPR) Workshops}, pages 2909--2918, 2020.
\newblock \doi{10.1109/CVPRW50498.2020.00348}.

\bibitem[Yu et~al.(2026)Yu, Liu, Zhao, Zhang, and Mei]{yu2026mfd}
Xuyi Yu, Yaohua Liu, Yinghai Zhao, Huipeng Zhang, and Kuizhi Mei.
\newblock Beyond single supervision: Multi-perspective quantization-aware
  distillation for semantic segmentation.
\newblock \emph{IEEE Transactions on Circuits and Systems for Video
  Technology}, 36\penalty0 (9):\penalty0 14168--14183, 2026.
\newblock \doi{10.1109/TCSVT.2026.3688538}.

\bibitem[Zhang and Chung(2021)]{zhang2021medq}
Rongzhao Zhang and Albert C.~S. Chung.
\newblock {MedQ}: Lossless ultra-low-bit neural network quantization for
  medical image segmentation.
\newblock \emph{Medical Image Analysis}, 73:\penalty0 102200, 2021.
\newblock \doi{10.1016/j.media.2021.102200}.

\bibitem[Zhao et~al.(2017)Zhao, Shi, Qi, Wang, and Jia]{zhao2017pspnet}
Hengshuang Zhao, Jianping Shi, Xiaojuan Qi, Xiaogang Wang, and Jiaya Jia.
\newblock Pyramid scene parsing network.
\newblock In \emph{Proceedings of the IEEE Conference on Computer Vision and
  Pattern Recognition (CVPR)}, pages 2881--2890, 2017.

\bibitem[Zhao and Zhao(2024)]{zhao2024sqakd}
Kaiqi Zhao and Ming Zhao.
\newblock Self-supervised quantization-aware knowledge distillation.
\newblock In \emph{Proceedings of the 27th International Conference on
  Artificial Intelligence and Statistics (AISTATS)}, volume 238 of
  \emph{Proceedings of Machine Learning Research}, pages 4375--4383, 2024.

\bibitem[Zhou et~al.(2016)Zhou, Wu, Ni, Zhou, Wen, and Zou]{zhou2016dorefa}
Shuchang Zhou, Yuxin Wu, Zekun Ni, Xinyu Zhou, He~Wen, and Yuheng Zou.
\newblock {DoReFa-Net}: Training low bitwidth convolutional neural networks
  with low bitwidth gradients, 2016.

\bibitem[Zhuang et~al.(2018)Zhuang, Shen, Tan, Liu, and Reid]{zhuang2018ptg}
Bohan Zhuang, Chunhua Shen, Mingkui Tan, Lingqiao Liu, and Ian Reid.
\newblock Towards effective low-bitwidth convolutional neural networks.
\newblock In \emph{Proceedings of the IEEE Conference on Computer Vision and
  Pattern Recognition (CVPR)}, pages 7920--7928, 2018.
\newblock arXiv:1711.00205.

\end{thebibliography}

\appendix
\titleformat{\section}{\normalfont\Large\bfseries}{Appendix \thesection.}{0.5em}{}
\makeatletter\renewcommand{\p@section}{Appendix~}\makeatother
\numberwithin{table}{section}\renewcommand{\thetable}{\Alph{section}.\arabic{table}}
\numberwithin{figure}{section}\renewcommand{\thefigure}{\Alph{section}.\arabic{figure}}
\section{Training setups}\label{app:setups}

Table~\ref{tab:data} lists the datasets and networks, Table~\ref{tab:setup-cls} gives the training setups of the classification results and Table~\ref{tab:setup-dense} those of the dense results; unless stated otherwise, a setup does not change across the networks of a dataset. The Tiny-ImageNet test images are the 10\,000 of its official validation set, CamVid has 11 classes plus void, and on NYUv2 the RMSE is computed over the pixels with valid depth. UNet-small has 32 and 64 channels in the encoder and a 128-channel bottleneck, and for depth the dense networks have a one-channel output head.

The classification teachers use SGD with momentum 0.9, and the classification students Adam for the weights and Adam without weight decay for the quantizer. The dense teachers use Adam without weight decay and batch 8, and the dense students one Adam optimizer with two parameter groups, without weight decay, with batch 8 and 120 epochs. Every learning rate follows a cosine decay, to zero for the students. The PTQ baseline uses one range per weight tensor in most layers; for classification it re-estimates the batch-normalization statistics on 49\,920 training images, and for the dense models it updates them on the 8 calibration batches.

A few settings change this setup. At W1A1, the CIFAR students use crops and flips without TrivialAugment and the one-batch initialization of the public code of SQAKD. One weight range per output channel is used where the ranges of the output channels differ most \cite{nagel2021whitepaper}: in the depthwise convolutions of MobileNetV2 and of the dense students, and in the linear layers of DeiT-Tiny, which also quantizes the attention softmax on a logarithmic grid. DeiT-Tiny at W8A8 sets the activation ranges at the 99.999th percentile and the weight ranges at the maximum of $|w|$. DeiT-Tiny at W2A2 uses learning rates of $5\cdot10^{-5}$ and $5\cdot10^{-7}$, no weight decay on LayerNorm, biases and tokens, ranges that minimize the mean squared error to the 2-bit grid, and 4-bit queries, keys and attention softmax. For VGG-13, the public code of SQAKD also keeps the second convolution of the first block in floating point \cite{zhao2024sqakd}, so our VGG-13 students quantize one layer more than theirs.

Most runs used the two NVIDIA L40S GPUs. The final CIFAR teachers, the Tiny-ImageNet teachers and some of the students ran on the two NVIDIA RTX A6000 GPUs: with the final teacher, some ResNet-20, VGG-8 and ResNet-32 runs, and on Tiny-ImageNet, seed~1; two MobileNetV2 QAT runs started there and finished on the L40S GPUs. The final CIFAR teachers use bfloat16 autocast in the forward pass.

At export, the per-tensor scale of each layer is repeated for each output channel, the format of ONNX Runtime static quantization, and the batch normalization that follows the layer is folded into these scales, so that each channel has its own scale; the integer codes of the weights are those of training.

The published results of Table~\ref{tab:cifar} come from the cited papers: EWGS from its paper, whose only CIFAR result with weights and activations at the same bit-width is ResNet-20 at W1A1 \cite{lee2021ewgs}; SQAKD, and the best of eleven distillation methods combined with EWGS, from Tables~4 and~7 of SQAKD, respectively \cite{zhao2024sqakd}; QKD and SPEQ from their papers \cite{kim2019qkd,boo2021speq}, where QKD distills from a ResNet-56 teacher (93.4\% on CIFAR-10, 73.4\% on CIFAR-100) trained jointly with the student and keeps the first and last layers at 8 bits; SQuaT from Tables~1 and~2 of SQuaT, which cover the ResNets only \cite{lee2026squat}; DA-QAKD from Table~II of DA-QAKD, which covers W2A2 and W4A4 \cite{kur2025daqakd}.

\begin{table*}[htbp]
\centering\footnotesize\setlength{\tabcolsep}{5pt}
\caption[Datasets and networks]{\textbf{Datasets and networks.} Test images, input resolution, networks with their number of parameters, and metric of each task.}
\label{tab:data}
\begin{tabular}{@{}L{0.16\linewidth}L{0.13\linewidth}L{0.11\linewidth}L{0.15\linewidth}L{0.28\linewidth}L{0.09\linewidth}@{}}
\toprule
Task & Dataset & Test images & Input & Networks (parameters) & Metric \\
\midrule
Classification & CIFAR-10 & 10\,000 & $32\times32$ & ResNet-20 (0.27\,M), VGG-8 (3.92\,M) & top-1 \\
 & CIFAR-100 & 10\,000 & $32\times32$ & ResNet-32 (0.47\,M), VGG-13 (9.46\,M) & top-1 \\
 & Tiny-ImageNet & 10\,000 & $64\times64$, upsampled to $224\times224$ & ResNet-18 (11.28\,M), MobileNetV2 (2.48\,M), DeiT-Tiny (5.60\,M) & top-1 \\
\addlinespace[2pt]
Segmentation & CamVid & 233 & $360\times480$ & R18-PSPNet (13.80\,M), Fast-SCNN (1.14\,M), UNet-small (0.47\,M) & mIoU \\
Depth estimation & NYUv2 & 654 & $288\times384$ & R18-PSPNet, Fast-SCNN, UNet-small & RMSE (m) \\
\bottomrule
\end{tabular}
\end{table*}

\begin{table*}[htbp]
\centering\footnotesize\setlength{\tabcolsep}{5pt}
\caption[Training setups for classification]{\textbf{Training setups for classification.} Hyper-parameters of the teachers and of the students on each dataset.}
\label{tab:setup-cls}
\begin{tabular}{@{}L{0.17\linewidth}L{0.24\linewidth}L{0.24\linewidth}L{0.27\linewidth}@{}}
\toprule
 & CIFAR-10 & CIFAR-100 & Tiny-ImageNet \\
Networks & ResNet-20, VGG-8 & ResNet-32, VGG-13 & ResNet-18, MobileNetV2, DeiT-Tiny \\
\midrule
\multicolumn{4}{@{}l}{\textbf{\Reft{} teachers}}\\
Learning rate & 0.1 & 0.05 & -- \\
Weight decay & $5\cdot10^{-4}$ & $5\cdot10^{-4}$ & -- \\
Batch size & 128 & 64 & -- \\
Epochs & 400 & 720 & -- \\
Augmentation & crop, flip & crop, flip & -- \\
\midrule
\multicolumn{4}{@{}l}{\textbf{Final teachers}}\\
Learning rate & 0.1 & 0.05 & 0.01 \\
Weight decay & $5\cdot10^{-4}$ & $5\cdot10^{-4}$ & $10^{-4}$ \\
Batch size & 128 & 64 & 128 \\
Epochs & 1200 (ResNet-20), 400 (VGG-8) & 720 & 30 \\
Augmentation & crop, flip, AutoAugment (CIFAR-10 policy), Cutout ($16\times16$) & crop, flip, AutoAugment (CIFAR-10 policy), Cutout ($16\times16$) & resized crop, color jitter, flip \\
Weight average & 0.9998 & 0.9999 & -- \\
\midrule
\multicolumn{4}{@{}l}{\textbf{Students}}\\
Learning rate & & & \\
\quad weights & $10^{-3}$ & $5\cdot10^{-4}$ & $5\cdot10^{-4}$ \\
\quad quantizer & $10^{-5}$ & $5\cdot10^{-6}$ & $5\cdot10^{-6}$ \\
Weight decay & $10^{-4}$ & $5\cdot10^{-4}$ & $5\cdot10^{-4}$ \\
Batch size & 256 & 64 & 64 \\
Epochs & 1200 (ResNet-20), 400 (VGG-8) & 720 & 100 \\
Augmentation & crop, flip, TrivialAugment & crop, flip, TrivialAugment & resized crop, color jitter, flip \\
BN statistics & \multicolumn{3}{l}{re-estimated on the training set before export} \\
\bottomrule
\end{tabular}
\end{table*}

\begin{table*}[htbp]
\centering\footnotesize\setlength{\tabcolsep}{5pt}
\caption[Training setups for segmentation and depth estimation]{\textbf{Training setups for segmentation and depth estimation.} Hyper-parameters of the teachers and of the students on each dataset. Three values in a cell follow the order Fast-SCNN, R18-PSPNet, UNet-small.}
\label{tab:setup-dense}
\begin{tabular}{@{}L{0.18\linewidth}L{0.38\linewidth}L{0.38\linewidth}@{}}
\toprule
 & Segmentation, CamVid & Depth estimation, NYUv2 \\
\midrule
Images & 367 training, 233 test & 795 training, 654 test \\
Resolution & training crops of $352\times352$, test $360\times480$ & $288\times384$ \\
Augmentation & scaling in $[0.75,1.5]$, crop, flip, photometric TrivialAugment & scaling in $[1.0,1.4]$, crop, flip, photometric TrivialAugment \\
Task loss & cross-entropy, void ignored & berHu, $c=0.2\max|d^s-d|$ \\
\midrule
\multicolumn{3}{@{}l}{\textbf{Teachers}}\\
Initialization & Cityscapes, Cityscapes, scratch & scratch, ImageNet-1k backbone, scratch \\
Learning rate & $10^{-3}$, $3\cdot10^{-4}$, $10^{-3}$ & $10^{-3}$, $3\cdot10^{-4}$, $10^{-3}$ \\
Epochs & 150 & 100, 100, 200 \\
\midrule
\multicolumn{3}{@{}l}{\textbf{Students}}\\
Learning rate & & \\
\quad weights & $3\cdot10^{-4}$ at W2A8, $10^{-4}$ at W4A8 and W8A8 & $3\cdot10^{-4}$ at W2A8, $10^{-4}$ at W4A8 and W8A8 \\
\quad quantizer & $10^{-5}$ & $10^{-5}$ \\
Pixel-wise term & KL, $T=4$ & L1 to the depth of the teacher \\
Pair-wise layer & pyramid pooling, pyramid pooling, bottleneck & pyramid pooling, pyramid pooling, bottleneck \\
Pair-wise $N$ & 121, 256, 256 & 81, 256, 256 \\
Holistic term & weight 0.1 & -- \\
Critic & four convolutions; Adam, learning rate $10^{-5}$, $\beta=(0.5,0.9)$ & -- \\
\bottomrule
\end{tabular}
\end{table*}

\section{Ablations}\label{app:ablations}

\subsection{Classification objective}\label{app:abl-cls}
Table~\ref{tab:abl-cls} compares the classification objective of Eq.~\eqref{eq:cls} with its two terms alone, the distillation term (the objective of SQAKD) and the cross-entropy, on CIFAR in the benchmark scheme. The students distill from a third teacher per network, the TrivialAugment teacher. It follows the \reft{} setup with TrivialAugment added, except on CIFAR-10, where it is trained with Adam (learning rate $10^{-3}$, batch size 256, weight decay $10^{-4}$) and, for ResNet-20, for 1200 epochs. The students also use TrivialAugment, with the training setup of \ref{app:setups}; every cell has three seeds except VGG-8 at W8A8 with CE+KL, which has one. The combined objective is above the distillation term alone in 10 of the 12 settings, and at most 0.02 points below it in the other two ($p < 0.005$, two-sided Wilcoxon signed-rank test over the settings). It has the highest mean in 6 of the 12 settings and is at most 0.64 points below the best of the three; the cross-entropy alone has the highest mean in 5, all at W8A8 or W4A4. Doubling the weight of the distillation term, as in DA-QAKD, does not improve on unit weights on average: over eight settings, with one to three seeds each, the change ranges from $-0.27$ to $+0.33$ points, with a mean of $-0.05$.

\begin{table*}[tbp]
\centering\scriptsize\setlength{\tabcolsep}{4pt}
\caption[Classification objective]{\textbf{Classification objective.} Top-1 accuracy (\%) of CIFAR students in the benchmark scheme with the distillation term alone (KL), the cross-entropy alone (CE) or both (CE+KL), mean and standard deviation over three seeds (one seed where no standard deviation is shown), with the FP32 teacher in parentheses. Bold: highest mean.}
\label{tab:abl-cls}
\begin{tabular}{@{}llccc@{}}
\toprule
Network (teacher) & Bits & KL & CE & CE+KL \\
\midrule
\multicolumn{5}{@{}l}{\textbf{CIFAR-10}}\\
ResNet-20 (93.37) & W8A8 & 93.51\pmsd{0.03} & \textbf{93.83\pmsd{0.17}} & 93.54\pmsd{0.03} \\
 & W4A4 & 93.18\pmsd{0.04} & \textbf{93.81\pmsd{0.15}} & 93.17\pmsd{0.03} \\
 & W2A2 & \textbf{91.86\pmsd{0.11}} & 91.62\pmsd{0.05} & 91.84\pmsd{0.18} \\
\addlinespace[2pt]
VGG-8 (92.58) & W8A8 & 92.59\pmsd{0.05} & \textbf{92.79\pmsd{0.15}} & 92.62 \\
 & W4A4 & 92.59\pmsd{0.08} & \textbf{92.84\pmsd{0.18}} & 92.81\pmsd{0.05} \\
 & W2A2 & 91.74\pmsd{0.10} & 91.43\pmsd{0.11} & \textbf{92.03\pmsd{0.02}} \\
\midrule
\multicolumn{5}{@{}l}{\textbf{CIFAR-100}}\\
ResNet-32 (74.49) & W8A8 & 74.67\pmsd{0.04} & \textbf{75.08\pmsd{0.41}} & 74.93\pmsd{0.03} \\
 & W4A4 & 74.17\pmsd{0.15} & 74.43\pmsd{0.23} & \textbf{74.49\pmsd{0.14}} \\
 & W2A2 & 70.01\pmsd{0.14} & 70.14\pmsd{0.04} & \textbf{70.83\pmsd{0.18}} \\
\addlinespace[2pt]
VGG-13 (77.49) & W8A8 & 77.91\pmsd{0.17} & 76.97\pmsd{0.13} & \textbf{78.29\pmsd{0.19}} \\
 & W4A4 & 77.73\pmsd{0.30} & 77.11\pmsd{0.17} & \textbf{78.37\pmsd{0.22}} \\
 & W2A2 & 76.18\pmsd{0.17} & 74.79\pmsd{0.10} & \textbf{76.53\pmsd{0.01}} \\
\bottomrule
\end{tabular}
\end{table*}

\subsection{Dense objectives}\label{app:abl-dense}
Table~\ref{tab:abl-dense} reports the segmentation objective at W2A2 in the benchmark scheme without the holistic or without the pair-wise term, with three seeds, for UNet-small on CamVid and for Fast-SCNN on Cityscapes \cite{cordts2016cityscapes}, which has 19 classes. The holistic term contributes most: removing it lowers the mIoU by 1.55 and 5.12 points, against 0.11 and 0.85 points for the pair-wise term.

\begin{table*}[tbp]
\centering\scriptsize\setlength{\tabcolsep}{4pt}
\caption[Dense objectives]{\textbf{Dense objectives.} mIoU (\%) of QAT and of the segmentation objective with all terms, without the holistic term or without the pair-wise term, at W2A2 in the benchmark scheme, mean over three seeds.}
\label{tab:abl-dense}
\begin{tabular}{@{}lcc@{}}
\toprule
Objective & UNet-small, CamVid & Fast-SCNN, Cityscapes \\
\midrule
QAT (task loss only) & 42.31 & 45.14 \\
Eq.~\eqref{eq:dense}, all terms & 43.38 & 49.21 \\
without the holistic term & 41.83 & 44.09 \\
without the pair-wise term & 43.27 & 48.36 \\
\bottomrule
\end{tabular}
\end{table*}

\section{Effect of a more accurate teacher}\label{app:teacher}
In Table~\ref{tab:teacher}, the students keep the same loss, quantizer and augmentation, and only the teacher changes, from the \reft{} to the final one; because the student starts from the teacher weights, the change affects both its initialization and its distillation target. From W2A2 to W8A8, the final teacher gives a better student in all 12 settings and in all three seeds of each, by 0.34 to 4.06 points (mean over seeds).

\begin{table*}[tbp]
\centering\scriptsize\setlength{\tabcolsep}{4pt}
\caption[Effect of a more accurate teacher]{\textbf{Effect of a more accurate teacher.} Top: top-1 accuracy (\%) of the FP32 teachers. Bottom: improvement of the \ours{} student with the final over the \reft{} teacher, paired by seed, mean and standard deviation over three seeds.}
\label{tab:teacher}
\begin{tabular}{@{}lcccc@{}}
\toprule
 & \multicolumn{2}{c}{CIFAR-10} & \multicolumn{2}{c}{CIFAR-100} \\
\cmidrule(lr){2-3}\cmidrule(lr){4-5}
 & ResNet-20 & VGG-8 & ResNet-32 & VGG-13 \\
\midrule
\multicolumn{5}{@{}l}{\textbf{FP32 teacher}}\\
\quad \reft{} & 92.59 & 92.18 & 71.69 & 76.81 \\
\quad final & 94.47 & 93.33 & 76.47 & 78.62 \\
\quad improvement & +1.88 & +1.15 & +4.78 & +1.81 \\
\addlinespace[2pt]
\multicolumn{5}{@{}l}{\textbf{Improvement of the \ours{} student}}\\
\quad W8A8 & +1.49\pmsd{0.03} & +0.80\pmsd{0.06} & +4.06\pmsd{0.11} & +1.23\pmsd{0.17} \\
\quad W4A4 & +1.49\pmsd{0.09} & +0.66\pmsd{0.05} & +3.54\pmsd{0.23} & +1.32\pmsd{0.09} \\
\quad W2A2 & +0.34\pmsd{0.07} & +0.40\pmsd{0.16} & +0.49\pmsd{0.29} & +0.88\pmsd{0.21} \\
\bottomrule
\end{tabular}
\end{table*}

\section{Paired comparison with QAT}\label{app:stats}

QAT and \ours{} share the student architecture, the starting weights, the training setup and the seeds, so each seed gives a paired difference. Table~\ref{tab:ci-qat} reports the mean of the three differences and its 95\% confidence interval (Student $t$, two degrees of freedom), in the device scheme on CIFAR, CamVid and NYUv2 and in the benchmark scheme on Tiny-ImageNet. Over the 18 dense settings, \ours{} is better than QAT in 43 of the 54 paired runs, with no ties; a two-sided sign test on these runs, treated as independent, gives $p = 1.4\times10^{-5}$. With the setting as the unit, which does not require independent runs, the sign test on the 15 of 18 settings with a better mean gives $p = 0.0075$.

\begin{table*}[tbp]
\centering\scriptsize\setlength{\tabcolsep}{4pt}
\caption[Paired comparison with QAT]{\textbf{Paired comparison with QAT.} Mean paired difference between \ours{} and QAT over three seeds, positive when \ours{} is better, with its 95\% confidence interval in square brackets and, in parentheses, the seeds in which \ours{} is better.}
\label{tab:ci-qat}
\begin{tabular}{@{}llll@{}}
\toprule
\textbf{CIFAR-10} & W8A8 & W4A8 & W2A8 \\
ResNet-20 & +0.31 [+0.02, +0.60] (3/3) & +0.20 [+0.07, +0.34] (3/3) & +0.36 [+0.21, +0.50] (3/3) \\
VGG-8 & +0.72 [+0.16, +1.28] (3/3) & +0.73 [+0.57, +0.89] (3/3) & +0.69 [+0.52, +0.86] (3/3) \\
\midrule
\textbf{CIFAR-100} & W8A8 & W4A8 & W2A8 \\
ResNet-32 & +1.69 [+1.26, +2.13] (3/3) & +1.66 [+0.82, +2.50] (3/3) & +1.78 [+1.72, +1.83] (3/3) \\
VGG-13 & +2.40 [+2.15, +2.64] (3/3) & +2.36 [+1.47, +3.25] (3/3) & +3.50 [+2.75, +4.25] (3/3) \\
\midrule
\textbf{Tiny-ImageNet} & W8A8 & W4A4 & W2A2 \\
ResNet-18 & +5.66 [+4.90, +6.42] (3/3) & +5.73 [+5.40, +6.06] (3/3) & +9.68 [+8.87, +10.49] (3/3) \\
MobileNetV2 & +4.49 [+3.88, +5.10] (3/3) & +4.55 [+3.28, +5.82] (3/3) & +17.45 [+12.63, +22.27] (3/3) \\
DeiT-Tiny & +10.37 [+9.87, +10.88] (3/3) & +14.15 [+13.42, +14.87] (3/3) & +2.19 [+1.38, +3.00] (3/3) \\
\midrule
\textbf{CamVid (mIoU)} & W8A8 & W4A8 & W2A8 \\
Fast-SCNN & +0.78 [+0.60, +0.97] (3/3) & +0.95 [$-$0.43, +2.33] (3/3) & +0.53 [$-$0.60, +1.66] (3/3) \\
R18-PSPNet & +0.22 [$-$0.27, +0.70] (3/3) & +0.22 [+0.05, +0.40] (3/3) & +1.16 [+0.42, +1.89] (3/3) \\
UNet-small & $-$0.56 [$-$1.16, +0.03] (0/3) & +0.04 [$-$0.68, +0.76] (2/3) & $-$1.10 [$-$3.63, +1.43] (0/3) \\
\midrule
\textbf{NYUv2 (RMSE, mm)} & W8A8 & W4A8 & W2A8 \\
Fast-SCNN & +2.6 [$-$4.0, +9.1] (2/3) & +4.0 [$-$4.1, +12.1] (3/3) & $-$4.4 [$-$17.5, +8.8] (1/3) \\
R18-PSPNet & +1.3 [$-$1.0, +3.7] (3/3) & +2.2 [$-$1.6, +6.0] (3/3) & +3.0 [$-$4.4, +10.5] (2/3) \\
UNet-small & +4.8 [+2.5, +7.0] (3/3) & +3.6 [+1.4, +5.8] (3/3) & +2.8 [$-$0.3, +6.0] (3/3) \\
\bottomrule
\end{tabular}
\end{table*}

\end{document}